\documentclass[10pt, journal]{IEEEtran}

\usepackage{times}
\usepackage{multicol}
\usepackage[bookmarks=true]{hyperref}
\usepackage{xcolor}
\usepackage{hyperref}
\usepackage{amsmath, amssymb}
\usepackage{amsfonts}
\usepackage{graphicx}
\usepackage{siunitx}
\usepackage{standalone}
\usepackage{booktabs}
\usepackage[ruled,vlined,linesnumbered]{algorithm2e}
\usepackage{mdframed}
\usepackage{fancyvrb}
\usepackage{soul}
\usepackage{dsfont,mathabx}
\usepackage{array, booktabs}
\usepackage{threeparttable}
\usepackage{multirow}

\usepackage{subfigure}
\usepackage{amsthm}
\newtheorem{theorem}{Theorem}
\newtheorem{lemma}[theorem]{Lemma}

\theoremstyle{definition}
\newtheorem{remark}{Remark}
\newtheorem{example}{Example}
\newtheorem{problem}{Problem}
\newtheorem{definition}{Definition}

\title{
Time Minimization and Online Synchronization\\ for Multi-agent Systems under \\
Collaborative Temporal Tasks
}

\author{Zesen Liu, Meng Guo and Zhongkui Li
\thanks{The authors are with the State Key Laboratory
    for Turbulence and Complex Systems,
    Department of Mechanics and Engineering Science,
    College of Engineering, Peking University, Beijing 100871, China.
    E-mail: \texttt{1901111653, meng.guo, zhongkli@pku.edu.cn}.
    This work was supported by the National Natural Science Foundation
    of China under grants 61973006, T2121002; and by Beijing Natural
    Science Foundation under grant JQ20025.
}
}

\begin{document}

\maketitle

\begin{abstract}
Multi-agent systems can be extremely efficient when solving a team-wide task in a concurrent manner.
However, without proper synchronization, the correctness of the combined behavior is hard to guarantee,
such as to follow a specific ordering of sub-tasks or to perform a simultaneous collaboration.
This work addresses the minimum-time task planning problem for multi-agent systems
under complex global tasks stated as Linear Temporal Logic (LTL) formulas.
These tasks include the temporal and spatial requirements on both independent local actions and direct sub-team collaborations.
The proposed solution is an anytime algorithm that combines the partial-ordering analysis
of the underlying task automaton for task decomposition,
and the branch and bound (BnB) search method for task assignment.
Analyses of its soundness, completeness and optimality as the minimal completion time are provided.
It is also shown that a feasible and near-optimal solution is quickly reached while the search continues within the time budget.
Furthermore, to handle fluctuations in task duration and agent failures during online execution,
an adaptation algorithm is proposed to synchronize execution status and re-assign unfinished subtasks dynamically to maintain correctness and optimality.
Both algorithms are validated rigorously over large-scale systems via numerical simulations and hardware experiments,
against several strong baselines.
\end{abstract}



\section{Introduction}\label{sec:introduction}
Multi-agent systems consist of a fleet of homogeneous or heterogeneous robots,
such as autonomous ground vehicles and aerial vehicles.
They are often deployed to accomplish tasks that are otherwise too inefficient or even infeasible for a single robot in~\cite{arai2002advances}.
First,
by allowing the robots to move and act concurrently,
the overall efficiency of the team can be significantly improved.
For instance,
a fleet of delivery vehicles can drastically reduce the delivery time~\cite{toth2002overview};
and a team of drones can surveil a large terrain and detect poachers~\cite{cliff2015online}.
Second,
by enabling multiple robots to directly collaborate on a task,
capabilities of the team can be greatly extended.
For instance,
several mobile manipulators can transport objects that are otherwise too heavy for one~\cite{fink2008multi};
and a team of mobile vehicles can collaboratively herd moving targets via formation~\cite{varava2017herding}.
Furthermore, to specify these complex tasks,
many recent work propose to use formal languages such as Linear Temporal Logic (LTL)
formulas ~\cite{baier2008principles}, as an intuitive yet powerful way to describe
both spatial and temporal requirements on the team behavior,
see~\cite{ulusoy2013optimality, kantaros2020stylus, schillinger2018simultaneous, guo2018multirobot} for examples.

\begin{table*}[t]
  \begin{center}
    \caption{Comparison of related work as discussed in Sec.~\ref{subsec:multi-ltl},
  regarding the problem formulation and key features.}
\label{table:compare}
\begin{adjustbox}{width=0.8\linewidth}
{\def\arraystretch{1.2}\tabcolsep=3pt
\begin{tabular}{cccccccc}
\toprule
Ref. & Syntax & Collaboration & Objective  & Solution & Anytime & Synchronization & Adaptation \\ \midrule
\cite{guo2015multi, tumova2016multi} & Local-LTL & No & Summed Cost & Dijkstra & No & Event-based  & Yes\\
\cite{guo2016task} & Local-LTL & Yes & Summed Cost & Dijkstra &  No & Event-based  & Yes \\
\cite{kantaros2020stylus, luo2021abstraction} & Global-LTL & No & Summed Cost & Sampling & Yes & All-time & No \\
\cite{schillinger2018simultaneous} & Global-LTL & No & Balanced & Martins' Alg.  & No & None & No \\
\cite{luo2021temporal} & Global-LTL & Yes & Summed Cost & MILP & No & All-time & No \\
\cite{sahin2019multirobot, jones2019scratchs} & Global-cLTL & No & Summed Cost & MILP & No & Partial & No \\
\textbf{Ours} & Global-LTL & Yes & Min Time & BnB & Yes & Event-based & Yes \\
\bottomrule
\end{tabular}%
}
\end{adjustbox}
\end{center}
\end{table*}

However, coordination of these robots to accomplish the desired task can result in great complexity,
as the set of possible task assignments can be combinatorial with respect to the number of robots
and the length of tasks~\cite{toth2002overview, LAVAEI2022110617}.
It is particularly so when certain metric should be minimized,
such as the completion time or the summed cost of all robots.
Considerable results are obtained in many recent work, e.g.,
via optimal planning algorithms
such as mixed integer linear programming (MILP) in~\cite{luo2021temporal,
 sahin2019multirobot, jones2019scratchs},
and search algorithms over state or solution space as
in~\cite{kantaros2020stylus, schillinger2018simultaneous, luo2021abstraction}.
Whereas being sound and optimal,
many existing solutions are designed from an offline perspective,
thus lacking in one of the following aspects that could be essential for robotic missions:
(I) \emph{Real-time requirements}.
Optimal solutions often take an unpredictable amount of time to compute,
without any intermediate feedback.
However, for many practical applications,
good solutions that are generated fast and reliably are often more beneficial;
(II) \emph{Synchronization} during plan execution.
Many of the derived plans are assigned to the robots
and executed independently without any synchronization among them,
e.g., no coordination on the start and finish time of a subtask.
Such procedure generally relies on a precise model of the underlying system
such as traveling time and task execution time,
and the assumption that no direct collaborations are required.
Thus, any mismatch in the given model or dependency in the subtasks would lead to erroneous execution;
(III) \emph{Online adaptation}.
An optimal but static solution can not handle changes in the workspace or in the team during
online execution, e.g., certain paths between subtasks are blocked, or some robots break down.

To overcome these challenges,
this work takes into account the minimum-time task coordination problem of a team of heterogeneous robots.
The team-wise task is given as LTL formulas over the desired actions to perform at the
desired regions of interest in the environment.
Such action can be independent that it can be done by one of these robots alone,
or collaborative where several robots should collaborate to accomplish it.
The objective is to find an optimal task policy for the team such that the completion time for the task is minimized.
Due to the NP-completeness of this problem,
the focus here is to design an anytime algorithm that returns the best solution within the given time budget.
More specifically,
the proposed algorithm interleaves between the partial ordering analysis
of the underlying task automaton for task decomposition,
and the branch and bound (BnB) search method for task assignment.
Each of these two sub-routines is anytime itself.
The proposed partial relations can be applied to non-instantaneous subtasks,
thus providing a more general model for analyzing concurrent subtasks.
Furthermore, the proposed lower and upper bounding methods during the BnB search
significantly reduces the search space.
The overall algorithm is proven to be complete and sound for the considered objective,
and also shown empirically that a feasible and near-optimal solution is quickly reached.
Besides,
an online synchronization protocol is proposed to handle fluctuations in the execution time,
while ensuring that the partial ordering constraints are still respected.
Lastly, to handle agent failures during the task execution,
an adaptation algorithm is proposed to dynamically reassign unfinished subtasks
online to maintain optimality.
The effectiveness and advantages of the proposed algorithm
are demonstrated rigorously via numerical simulations and hardware experiments,
particularly over {large-scale} systems of more than $20$ robots and $10$ subtasks.

The main contribution of this work is threefold:
(I) it extends the existing work on temporal task planning to allow collaborative subtasks
and the practical objective of minimizing task completion time;
(II) it proposes an anytime algorithm that is sound, complete and optimal,
which is particularly suitable for real-time applications where computation time is restricted;
and (III) it provides a novel theoretical approach that combines the partial ordering
analysis for task decomposition and the BnB search for task assignment.

The rest of the paper is organized as follows:
Sec.~\ref{sec:related-work} reviews related work.
The formal problem description is given in Sec.~\ref{sec:problem}.
Main components of the proposed framework are presented in
Sec.~\ref{subsec:initial-synthesis}-\ref{subsec:summary}.
Experiment studies are shown in Sec.~\ref{sec:experiments},
followed by conclusions and future work in Sec.~\ref{sec:conclusion}.

\section{Related Work}\label{sec:related-work}

\subsection{Multi-agent Task and Motion Planning}\label{subsec:multi-tamp}
Planning for multi-agent systems have two distinctive characteristics:
high-level task planning in the discrete task space,
and low-level motion planing in the continuous state space.
In particular,
given a team-wise task, task planning refers to the process of first decomposing this task into sub-tasks
and then assigning them to the team, see~\cite{torreno2017cooperative,gini2017multi, khamis2015multi} for comprehensive surveys.
Such tasks can have additional constraints,
such as time windows~\cite{luo2015distributed}, robot capacities~\cite{fukasawa2006robust},
and ordering constraints~\cite{boerkoel2013distributed, nunes2015multi}.
The optimization criteria can be single or multiple,
two of which are the most common:
MinSUM that minimizes the sum of robot costs over all robots~\cite{gini2017multi, luo2015distributed, fukasawa2006robust},
and MinMAX that minimizes the maximum cost of a robot over all robots~\cite{nunes2015multi},
similar to the \emph{makespan} of all tasks.
Typical solutions can be categorized into centralized methods such as
Mixed Integer Linear Programming (MILP)~\cite{torreno2017cooperative} and search-based methods~\cite{fukasawa2006robust,FANG2022110228};
and decentralized methods such as
market-based methods~\cite{luo2015distributed}, distributed constraint optimization (DCOP) \cite{boerkoel2013distributed}.
However, since many task planning problems are in general NP-hard or even NP-complete~\cite{gini2017multi},
meta-heuristic approaches are used to gain computational efficiency,
e.g., local search~\cite{hoos2004stochastic} and genetic algorithms \cite{khamis2015multi}.
One type of problem is particularly of relevance to this work,
namely the Multi-Vehicle Routing Problem (MVRP) in operation research~\cite{gini2017multi, khamis2015multi},
where a team of vehicles are deployed to provide services at a set of locations,
and the MinSum objective above is optimized.
Despite of the similarity,
this work considers a significantly more general specification of team-wise tasks,
which can include as special cases the vanilla formulation and its variants with
temporal and spatial constraints~\cite{boerkoel2013distributed, nunes2015multi}.
Moreover, collaborative tasks and synchronization during online execution are often neglected in the aforementioned work, where planning and execution are mostly decoupled.

On the other hand,
motion planning problem for multi-agent systems aims at designing cooperative control policies such that a team-wise control objective is reached,
e.g., collision-free navigation~\cite{lavalle2006planning}, formation~\cite{chen2005formation},
consensus~\cite{li2009consensus} and coverage \cite{mesbahi2010graph}.
Such objectives are self-sustained but lacks a high-level purpose for task completion.
In this work, we extend these results by incorporating them into the team-wise task as collaborative sub-tasks.

\subsection{Temporal Logic Tasks}\label{subsec:multi-ltl}

\begin{figure}[t!]
	\centering
	\includegraphics[width=0.95\linewidth]{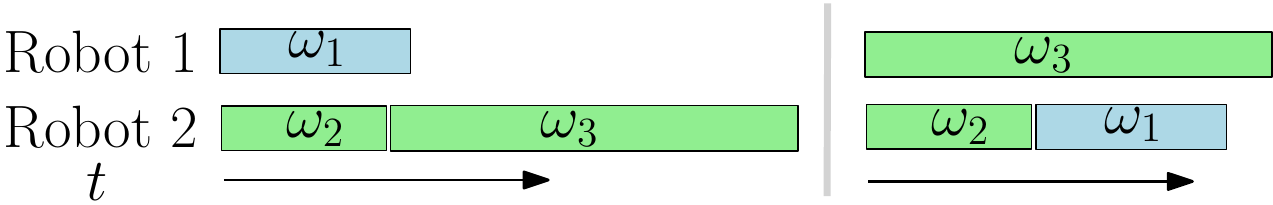}
	\caption{Comparison of the planning results based on decompositional states
		in~\cite{schillinger2018simultaneous} (\textbf{left}) and the partial ordering proposed in this work (\textbf{right}).
		Note that the sub-task~$\omega_3$ has to be {completed}
                after~$\omega_2$, while~$\omega_1$ is independent.}
		\label{fig:concurrent}
\end{figure}

Temporal logic formulas can be used to specify complex robotic tasks,
such as Probabilistic Computation Tree Logic (PCTL) in~\cite{lahijanian2011temporal},
Linear Temporal Logics (LTL) in~\cite{kantaros2020stylus, schillinger2018simultaneous,
 guo2015multi, chen2011formal},
and counting LTL (cLTL) in~\cite{sahin2019multirobot}.
As summarized in Table~\ref{table:compare},
the most related work can be compared in the following four aspects:
(i) \emph{collaborative tasks}.
In a bottom-up fashion,~\cite{guo2015multi, tumova2016multi, guo2016task} assume local LTL tasks and dynamic environment,
where collaborative tasks are allowed in~\cite{guo2016task}.
In a top-down fashion,~\cite{kantaros2020stylus, schillinger2018simultaneous, luo2021temporal, sahin2019multirobot, jones2019scratchs} consider team-wise tasks,
but no direct collaboration among the agents;
(ii) the \emph{optimization criteria}.
Most aforementioned work~\cite{kantaros2020stylus, guo2016task, luo2021temporal, sahin2019multirobot, jones2019scratchs} optimizes
the summed cost of all agents,
while~\cite{schillinger2018simultaneous} evaluates a weighted balance between this cost and
the task completion time. 
Even though both objectives are valid,
we emphasize in this work the achievement of maximum efficiency by minimizing solely the completion time;
(iii) the \emph{synchronization requirement}.
Synchronization happens when two or more agents communicate regarding the starting time of next the next sub-task.
The work in~\cite{kantaros2020stylus, luo2021abstraction, sahin2019multirobot} requires
full synchronization before each sub-task due to the product-based solution,
while~\cite{schillinger2018simultaneous} imposes no synchronization by allowing
only independent sub-tasks and thus limiting efficiency.
This work however proposes an online synchronization strategy for sub-tasks
that satisfies both the strict partial ordering and the simultaneous collaboration.
As illustrated in Fig.~\ref{fig:concurrent},
this can improve greatly the concurrency and thus the efficiency of the multi-agent execution even further;
and lastly
(iv) the \emph{solution basis}.
Solutions based on solving a MILP as in~\cite{luo2021temporal, sahin2019multirobot, jones2019scratchs} often can not guarantee a feasible solution within a time budget,
while this work proposes an anytime algorithm that could return quickly a feasible
and near-optimal solution.
Last but not least,
instead of generating only a static team-wise plan as in the aforementioned work,
the proposed online adaptation algorithm can handle fluctuations in the task duration
and possible agent failures during execution.

\subsection{Branch and Bound}\label{subsec:BnB-search}
Branch and bound (BnB) is a search paradigm to solve discrete and combinatorial optimization problems exactly~\cite{lawler1966branch, morrison2016branch}.
All candidate solutions are represented by a rooted tree within the solution space.
Via branching and bounding intelligently,
an efficient search strategy can be derived by pruning early branches that are provably
suboptimal. It has been successfully applied to various NP-hard problems, e.g.,
the traveling salesman problem~\cite{junger1995traveling},
multi-vehicle routing problem~\cite{gini2017multi, khamis2015multi},
and the job-shop scheduling problem~\cite{brucker1994branch}.
Different from the commonly-seen BnB algorithms that are applied to MILP directly,
we propose in this work a novel BnB search strategy \emph{specifically} designed
for the planning problem of multi-agent systems under complex temporal tasks.
It takes into account both the partial ordering constraints imposed by the temporal task,
and the synchronization constraints due to inter-agent collaboration.

\section{Preliminary}\label{sec:preliminary}
This section contains the preliminaries essential for this work,
namely, Linear Temporal Logic, B\"uchi Automaton,
and the general definition of partially ordered set,
\subsection{Linear Temporal Logic (LTL)}\label{subsec:LTL}
Linear Temporal Logic (LTL) formulas are composed of a set of atomic propositions $AP$
in addition to several Boolean and temporal operators. Atomic propositions are Boolean variables that can
be either true or false.  Particularly, the syntax~\cite{baier2008principles} is given as follows:
$\varphi \triangleq \top \;|\; p  \;|\; \varphi_1 \wedge \varphi_2  \;|\; \neg \varphi  \;|\; \bigcirc \varphi  \;|\;  \varphi_1 \,\textsf{U}\, \varphi_2,$
where $\top\triangleq \texttt{True}$, $p \in AP$, $\bigcirc$ (\emph{next}),
$\textsf{U}$ (\emph{until}) and $\bot\triangleq \neg \top$.
For brevity, we omit the derivations of other operators like $\Box$ (\emph{always}),
 $\Diamond$ (\emph{eventually}), $\Rightarrow$ (\emph{implication}).
The full semantics and syntax of LTL are omitted here for brevity,
see e.g.,~\cite{baier2008principles}.

An infinite {word} $w$ over the alphabet $2^{AP}$ is defined as an
infinite sequence ${\omega}=\sigma_1\sigma_2\cdots, \sigma_i\in 2^{AP}$.
The language of $\varphi$ is defined as the set of words that satisfy $\varphi$,
namely, $L_\varphi=Words(\varphi)=\{\boldsymbol{\omega}\,|\,\omega\models\varphi\}$
and $\models$ is the satisfaction relation.
However, there is a special class of LTL formula called \emph{co-safe} formulas,
which can be satisfied by a set of finite sequence of words.
They only contain the temporal operators $\bigcirc$, $\textsf{U}$ and $\Diamond$
 and are written in positive normal form.

\subsection{Nondeterministic B\"uchi Automaton}\label{subsec:nba}

Given an LTL formula $\varphi$ mentioned above, the associated Nondeterministic B\"{u}chi Automaton (NBA) can be derived with the following structure.
\begin{definition}[NBA] \label{def:nba}
A NBA $\mathcal{B}$ is a 5-tuple: $\mathcal{B}=(Q,\,Q_0,\,\Sigma,\,\delta,\,Q_F)$,
where $Q$ is the set of states;
$Q_0\subseteq Q$ is the set of initial states;
$\Sigma=AP$ is the allowed alphabet;
$\delta:Q\times \Sigma\rightarrow2^{Q}$ is the transition relation;
$Q_F\subseteq Q$ is the set of \emph{accepting} states. \hfill $\blacksquare$
\end{definition}

Given an infinite word $w=\sigma_1\sigma_2\cdots$, the resulting \emph{run}~\cite{baier2008principles} within $\mathcal{B}$
is an infinite sequence $\rho=q_0q_1q_2\cdots$
such that $q_0\in Q_0$, and $q_i\in Q$, $q_{i+1}\in\delta(q_i,\,\sigma_i)$ hold for all index~$i\geq 0$.
A run is called \emph{accepting} if it holds that
$\inf(\rho)\bigcap {Q}_F \neq \emptyset$,
where $\inf(\rho)$ is the set of states that appear in $\rho$ infinitely often.
In general, an accepting run can be written in the prefix-suffix structure,
where the prefix starts from an initial state and ends in an accepting state;
and the suffix is a cyclic path that contains the same accepting state.
It is easy to show that such a run is accepting if the prefix is repeated once,
while the suffix is repeated infinitely often.
In general, the size of~$\mathcal{B}$ is double exponential to the size of~$|\varphi|$.

\subsection{Partially Ordered Set}\label{subsec:partial}
A partial order~\cite{simovici2008mathematical} defined on a set $S$ is a relation $\rho\subseteq S\times S$,
which is reflexive, antisymmetric, and transitive.
Then, the pair $(S,\, \rho)$ is referred to as a partially ordered set (or simply \emph{poset}).
A generic partial order relation is given by $\leq$.
Namely, for $x,\,y\in S$, $(x,\,y)\in \rho$ if $x\leq y$.
The set $S$ is totally ordered if $\forall x,\,y\in S$, either $x\leq y$ or $y\leq x$ holds.
Two elements $x,\,y$ are incomparable if neither $x\leq y$ nor $y\leq x$ holds, denoted by $x\parallel y$.

\section{Problem Formulation}\label{sec:problem}

\begin{figure*}[t!]
	\centering
	\includegraphics[width=0.9\linewidth]{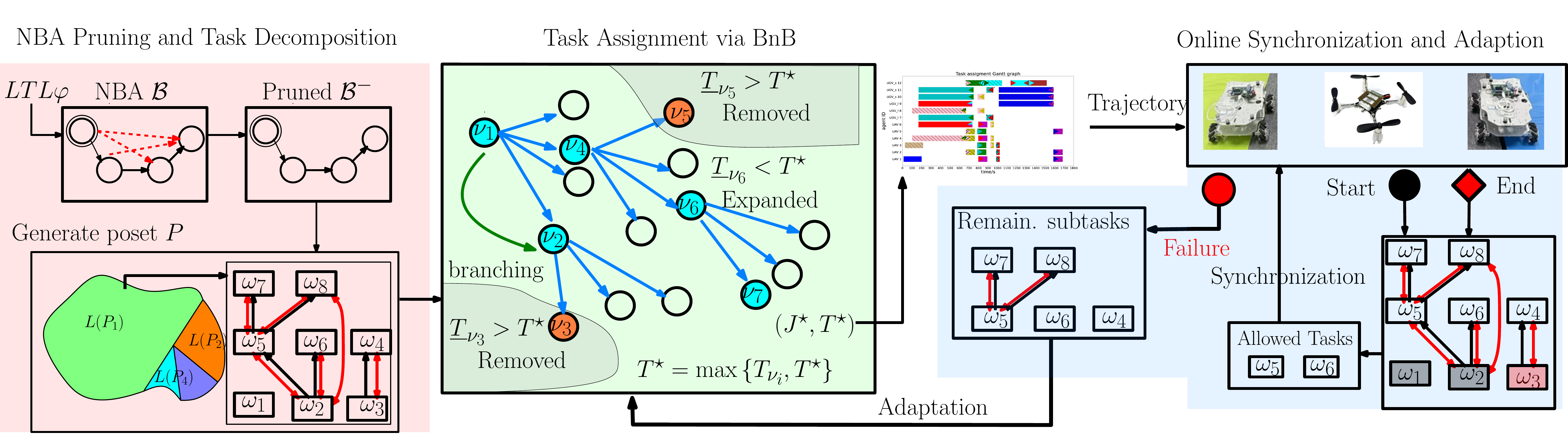}
	\caption{Overall structure of the proposed framework,
          which consists of three main parts:
        the computation of the posets, BnB search, and online execution.}
	\label{fig:logic_graph}
\end{figure*}

\subsection{Collaborative Multi-agent Systems}\label{subsec:multi-agent}

Consider a team of $N$ agents operating in a workspace as~${W}\subset \mathbb{R}^3$.
Within the workspace, there are $M$ regions of interest,
denoted by ${\mathcal{W}}\triangleq \{{W}_1,{W}_2,\,\cdots,{W}_M\}$, 
where~${W}_m\in {W}$.
Each agent~$n\in\mathcal{N}=\{1,\cdots,N\}$ can navigate within these regions
following its own transition graph, i.e., $\mathcal{G}_n=({\mathcal{W}},\,\rightarrow_n,\,d_n)$,
where $\rightarrow_n\subseteq {\mathcal{W}}\times {\mathcal{W}}$
is the allowed transition for agent~$n$;
and $d_n:\rightarrow_n \rightarrow \mathbb{R}_{+}$ maps each transition to its time duration.

Moreover, similar to the action model used in our previous work~\cite{guo2016task},
each robot $n\in \mathcal{N}$ is capable of performing a set of actions:
$\mathcal{A}_n\triangleq \mathcal{A}^{\texttt{l}}_n \cup \mathcal{A}^{\texttt{c}}_n$,
where $\mathcal{A}^{\texttt{l}}_n$ is a set of \emph{local} actions that can be independently performed by agent $n$ itself;
$\mathcal{A}^{\texttt{c}}_n$ is a set of \emph{collaborative} actions that can only be performed
in collaboration with other agents.
Moreover, denote by~$\mathcal{A}^c\triangleq\bigcup_{n\in\mathcal{N}}\mathcal{A}^c_n$,
$\mathcal{A}^l\triangleq\bigcup_{n\in\mathcal{N}}\mathcal{A}^l_n$.
More specifically,
there is a set of collaborative behaviors pre-designed for the team,
denoted by~$\mathcal{C}\triangleq \{C_1,\cdots, C_K\}$. 
Each behavior $C_k\in \mathcal{C}$ consists of a set of collaborative actions
that should be accomplished by different agents, namely:
\begin{equation}\label{eq:c-k}
C_k\triangleq \{a_1,\,a_2,\cdots,a_{\ell_k}\},
\end{equation}
where~$\ell_k>0$ is the number of collaborative actions required;
~$a_{\ell}\in \mathcal{A}^c$, $\forall \ell=1,\cdots,\ell_k$.
Thus, each collaborative action has a fixed duration as~$d:\mathcal{C}\rightarrow \mathbb{R}_{+}$.
For instance, formation~\cite{chen2005formation}, herding~\cite{pan2007multi},
 consensus~\cite{li2009consensus} and coverage~\cite{mesbahi2010graph}
 are common collaborative behaviors of multi-agent systems.

\begin{remark}\label{rm:collaborative}
Different from~\cite{luo2021temporal,sahin2019multirobot, jones2019scratchs},
the definition of collaborative behavior above does not specify explicitly
the agent identities or {types}, rather their capabilities.
This subtle difference can improve the flexibility of the underlying solution,
e.g.,  no hard-coding of the agent identities or types is necessary;
any capable agent can be recruited for the collaborative behavior.
\hfill $\blacksquare$
\end{remark}

\subsection{Task Specification}\label{subsec:task-specification}
First, the following three types of atomic propositions can be introduced:
(i) $p_m$ is \emph{true} when \emph{any} agent $n\in \mathcal{N}$ is at region ${W}_m\in {\mathcal{W}}$;
Let $\mathbf{p}\triangleq \{p_m,\, \forall {W}_m \in {\mathcal{W}}\}$;
(ii) $a^m_k$ is \emph{true} when a \emph{local} action~$a_k$ is performed at region ${W}_m\in{\mathcal{W}}$ by
agent~$n$, where $a_k \in \mathcal{A}_n^{\texttt{l}}$.
Let~$\mathbf{a} \triangleq\{a^m_k,\forall {W}_m \in \mathcal{W}, a_k\in \mathcal{A}^{l}\}$;
(iii) $c^m_k$ is \emph{true} when the collaborative behavior~$C_k$ in~\eqref{eq:c-k} is performed at region $W_m$.
Let $\mathbf{c} \triangleq\{c^m_k,\forall C_k \in \mathcal{C},\forall {W}_m \in {\mathcal{W}}\}$.

Given these propositions, a team-wide task specification can be specified as a sc-LTL formula
\begin{equation}\label{eq:task}
\varphi=\textup{sc-LTL}(\{\mathbf{p}, \mathbf{a}, \mathbf{c}\}),
\end{equation}
where the syntax of sc-LTL is introduced in Sec.~\ref{subsec:LTL}.
Denote by~$t_0,\, t_f>0$ the time instants when the system starts
and satisfies executing~$\varphi$, respectively.
Thus, the total time taken for the multi-agent team to satisfy~$\varphi$ is given by:
\begin{equation}\label{eq:task-freq}
T_\varphi = t_f-t_0.
\end{equation}
Since a sc-LTL can be satisfied in finite time, this total duration is finite and thus can be optimized.

\begin{remark}\label{rm:cost}
  As described previously in Sec.~\ref{sec:introduction}, the minimum time cost in~\eqref{eq:task-freq} is significantly different from the summed time cost of all agents, i.e.,
$\sum_{i} T_i$, where~$T_i$ is the total time agent~$i$ spent on executing task~$\varphi$,
as in~\cite{kantaros2020stylus, schillinger2018simultaneous, guo2016task, luo2021temporal}.
The main advantage of concurrent execution can be amplified via
the objective of time minimization,
since concurrent or sequential execution of the same task assignment could
be equivalent in terms of summed cost.
\hfill $\blacksquare$
\end{remark}
\begin{example}\label{exp:task}
Consider a team of UAVs and UGVs deployed for maintaining a remote Photovoltaic power plant.
One collaborative task considered in Sec.~\ref{sec:experiments} is given by:
\begin{equation}\footnotesize
\label{example:task} 
\begin{aligned}
\varphi_1 = & \Diamond(\texttt{repair}_{\texttt{p}_{3}} \wedge \lnot \texttt{scan}_{\texttt{p}_3} \wedge\Diamond \texttt{scan}_{\texttt{p}_3})
\wedge \Diamond (\texttt{wash}_{\texttt{p}_{21}} \wedge \\
&\Diamond \texttt{mow}_{\texttt{p}_{21}} \wedge \Diamond \texttt{scan}_{\texttt{p}_{21}}) \wedge \Diamond ( \texttt{sweep}_{\texttt{p}_{21}} \wedge \lnot \texttt{wash}_{\texttt{p}_{21}} \wedge\\
& \Diamond \texttt{mow}_{\texttt{p}_{21}}) \wedge \Diamond(\texttt{fix}_{\texttt{t}_5} \wedge \lnot \texttt{p}_{\texttt{18}}) \wedge \lnot \texttt{p}_{24} \,U \, \texttt{sweep}_{\texttt{p}_{27}} \\
&\wedge \Diamond (\texttt{wash}_{\texttt{p}_{34}} \wedge \bigcirc \texttt{scan}_{\texttt{p}_{34}}),
\end{aligned}
\end{equation}
which means to repair and scan certain PV panel $p_3$
in a given order, deeply cleaning the panel $p_{21},p_{34}$, fix transformers
$t_5$, sweep $p_{27}$ with safety request. 
 \hfill $\blacksquare$
\end{example}
\subsection{Problem Statement}\label{subsec:problem-statement}

\begin{problem}\label{prob:formulation}
Given the task specification~$\varphi$,
synthesize the motion and action
sequence for each agent $n\in \mathcal{N}$,
such that~$T_{\varphi}$ in~\eqref{eq:task-freq} is minimized.
\hfill $\blacksquare$
\end{problem}
Even through the above problem formulation is straightforward,
it is can be shown that this problem belongs to the class of NP-hard
problems~\cite{hochba1997approximation, bovet1994introduction},
as its core coincides with the makespan minimization problem
of flow-shop scheduling problems.
Various approximate algorithms have been proposed in~\cite{khamis2015multi, hoos2004stochastic}.
However,
the combination of dynamic vehicles within a graph-like environment,
linear temporal constraints,
and collaborative tasks has not been addressed.


In the following sections, we present the main solution of this work.
As shown in Fig.~\ref{fig:logic_graph}, the optimal plan synthesis which is performed offline is first
described in Sec.~\ref{subsec:initial-synthesis},
and then the online adaptation strategy to handle dynamic changes in Sec.~\ref{subsec:online-adaptation}.
The overall solution is summarized in Sec.~\ref{subsec:summary} with analyses
for completeness, optimality and complexity.

\section{Optimal Plan Synthesis}\label{subsec:initial-synthesis}
The optimal plan synthesis aims to solve Problem~\ref{prob:formulation} offline.
As mentioned previously in Sec.~\ref{subsec:multi-ltl},
most related work requires the synchronized product of all agents' model,
thus subject to exponential complexity.
Instead, we propose an anytime algorithm that combines seamlessly the partial-ordering analyses
of the underlying task automaton for task decomposition,
and the branch and bound (BnB) search method for task assignment.
In particular, it consists of three main components:
(i) the pre-processing of the NBA associated with the global task;
(ii) the partial-ordering analyses of the processed task automaton;
(iii) the BnB search algorithm that searches in the plan space
given the partial ordering constraints.

\subsection{B\"{u}chi Automaton Pruning}
\label{subsubsec:NBA-pruning}
As the first step, the NBA~$\mathcal{B}_{\varphi}$ associated with the task~$\varphi$
is derived, e.g., via translation tools in~\cite{gastin2001fast}.
Note that~$\mathcal{B}_{\varphi}$ has the structure as defined in Def.~\ref{def:nba},
which however can be overly redundant.
For instance, the required input alphabets for some transitions are infeasible for the whole team;
or some transitions are redundant as they can be decomposed equivalently
into other transitions.
Via detecting and removing such transitions, the size of the underlying NBA
can be greatly reduced, thus improving the efficiency of subsequent steps.
More specifically, pruning of~$\mathcal{B}_{\varphi}$ consists of the following three steps:

(i) Remove infeasible transitions.
Given any transition $q_j \in \delta(q_i,\, \sigma)$ in~$\mathcal{B}_{\varphi}$,
it is infeasible for the considered system if
no subgroup of agents in~$\mathcal{N}$ can generate $\sigma$.
It can be easily verified by checking whether there exist an agent that can navigate to region~$W_m$ and perform local action~$a_k$;
or several agents that can \emph{all} navigate to region~$W_m$ and perform collaborative action~$a_k$.
If infeasible, such transition is removed in the pruned automaton.

(ii) Remove invalid states.
Any state~$q\in Q$ in~$\mathcal{B}_{\varphi}$ is called invalid
if it can not be reached from any initial state;
or it can not reach any accepting state that is in turn reachable from itself.
An invalid state can not be part of an accepting path thus removed in the pruned automaton.

\begin{figure}
\centering
\includegraphics[scale=0.4]{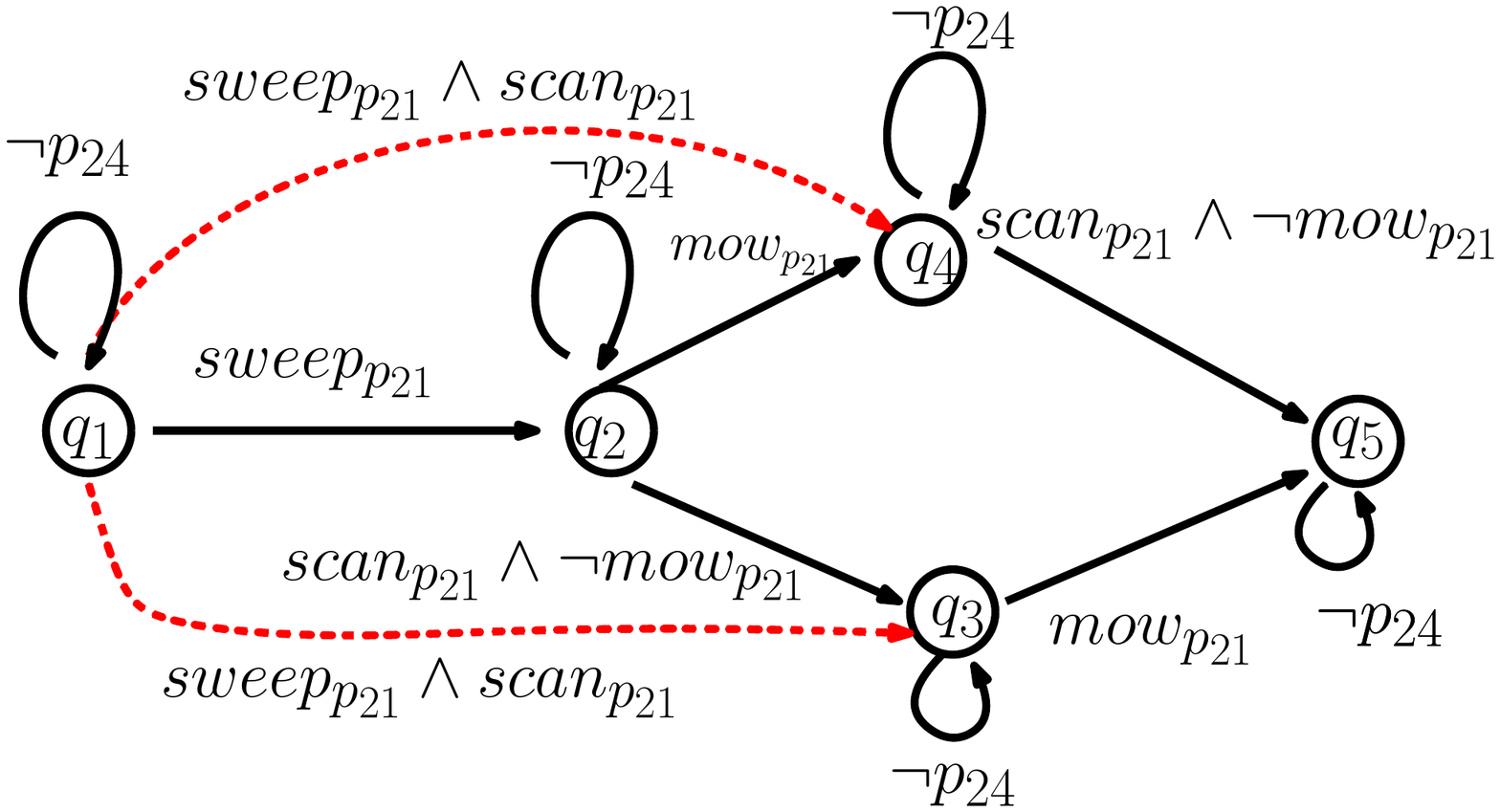}
\caption{
Example of decomposable transitions.
Any transition in red dashed line are all 
decomposable satisfied by Def.~\ref{def:decomposable-transition}.}
\label{fig:example_decomposable}
\end{figure}

(iii) Remove decomposable transitions.
Due to the distributed nature of multi-agent systems,
it is unrealistic to enforce the fulfillment of two or more subtasks
to be \emph{exactly} at the same instant in real time.
Therefore, if possible,
any transition that requires the simultaneous satisfaction of several sub-tasks is decomposed into equivalent transitions.
Decomposable transitions are formally defined in
Def.~\ref{def:decomposable-transition} below.
In other words, the input alphabets of a decomposed transition can be mapped to
two other transitions that connect the \emph{same} pair of states,
but via another intermediate state, as illustrated in Fig.~\ref{fig:example_decomposable}.
Thus, all decomposable transitions in~$\mathcal{B}_{\varphi}$ are removed in the pruned automaton.
Algorithmically, decomposability can be checked by simply composing and comparing the
propositional formulas associated with each transition.

\begin{definition}[Decomposable Transition]\label{def:decomposable-transition}
Any transition from state~$q_i$ to $q_j$ in~$\mathcal{B}_{\varphi}$ is decomposable if there exists another state $q_k$ such that $q_j\in \delta(q_i,\sigma_{ik}\cup\sigma_{kj})$ holds,
$\forall \sigma_{ik},\,\sigma_{kj} \subseteq \Sigma$ that $q_k\in \delta(q_i,\sigma_{ik})$ and $q_j\in \delta(q_k,\sigma_{kj})$ hold.
\hfill $\blacksquare$
\end{definition}

An example of decomposable transitions is shown in Fig.~\ref{fig:example_decomposable}.
To summarize, the pruned NBA, denoted by~$\mathcal{B}^{-}_{\varphi}$,
has the same structure as~$\mathcal{B}_{\varphi}$ but with much fewer states and edges.
In our experience, this pruning step can reduce up to $60\%$ states and edges for typical multi-agent tasks.
More details can be found in the experiment section. 
\begin{lemma}
  If there exists a word~$\omega$ that is accepted by~$\mathcal{B}$,
  then an equivalent run~$\omega'$ can be found that is accepted by~$\mathcal{B}^-$.
\end{lemma}
\begin{proof}
  Consider an accepting word $\omega=\cdots\{\sigma_1\land\sigma_2\}\cdots$.
  The associated run in~$\mathcal{B}$ is given by~$\rho=\cdots q_1 q_3\cdots$.
  Thus there exists state~$q_2$ satisfying that 
  $q_2=\delta(\sigma_1,q_1)$, $q_3=\delta(\sigma_2,q_2)$, $q_3=\delta(\sigma_1\land\sigma_2,q_1)$.
  Moreover, as the labels are not instantaneous, 
  an equivalent \emph{word} $\omega'=\cdots\{\sigma_1\land\sigma_2\}\{\sigma_1\land\sigma_2\}\cdots$
  can be created by adding an additional time step.
  The associated run in~$\mathcal{B}^-$ is given by~$\rho'=\cdots q_1 q_2 q_3\cdots$ where
  $q_2=\delta(\sigma_1\land\sigma_2,q_1)$, $q_3=\delta(\sigma_1\land\sigma_2,q_2)$.
  This completes the proof.
\end{proof}


\begin{example}\label{eq:prune-nba}
The NBA~$\mathcal{B}_{\varphi}$ associated with the task formula in~\eqref{example:task}
has $707$ states and $16044$ edges with more than $1.29\times10^7$ accepting \emph{word},
translated via~\cite{gastin2001fast}.
After the pruning process described above, the pruned automaton~$\mathcal{B}^{-}_{\varphi}$
has $707$ states and $2423$ edges with $174469$ accepting \emph{word}. The NBA pruning step reduces $84.9\%$ edges and $98.6\%$
accepting \emph{word} without losing any information.
\hfill $\blacksquare$
\end{example}

\subsection{Task Decomposition}\label{subsubsec:task-decompose}
Since the team-wise task in this work is given as a compact temporal task formula,
a prerequisite for optimal task assignment later is to decompose this task into suitable subtasks.
Moreover, different from simple reachability task,
temporal tasks can impose strict constraints on the ordering of sub-tasks.
For instance, the task~$\Diamond (\omega_1 \land \Diamond (\omega_2 \land \Diamond\omega_3)) $
specifies that $\omega_1,\omega_2,\omega_3$ should be satisfied in sequence,
while~$\Diamond\omega_1\land\Diamond\omega_2\land\Diamond\omega_3$ does not impose any ordering constraint.
Thus, it is essential for the overall correctness to abstract such ordering constraints among the subtasks.
This part describes how the subtasks and their partial orderings are abstracted from the pruned automaton~$\mathcal{B}_{\varphi}^{-}$.

\begin{definition}[Decomposition and Subtasks] \label{def:subtasks}
Consider an accepting run $\rho=q_0q_1\cdots q_L$ of~$\mathcal{B}_{\varphi}^{-}$,
where $q_0\in Q_0$ and $q_L\in Q_F$.
One \emph{possible} decomposition of $\varphi$ into subtasks is defined
as a set of 3-tuples:
\begin{equation}\label{eq:subtask}
\Omega_{\varphi} = \left\{(\ell,\, \sigma_\ell,\, \sigma^{s}_\ell),\, \forall \ell=1,\cdots,L\right\},
\end{equation}
where $\ell$ is the index of subtask~$\sigma_\ell$,
and $\sigma_\ell\subseteq \Sigma$ satisfies two conditions:
(i) $q_{\ell+1} \in \delta(q_{\ell-1},\,\sigma_\ell)$;
and (ii) $q_{\ell+1} \notin \delta(q_\ell,\,\sigma^-_\ell)$, where~$\sigma^-_\ell =  \sigma_\ell \backslash \{s\}$, $\forall s\in \sigma_\ell$, 
and $\sigma^{s}_\ell\subseteq \Sigma$ also satisfies two conditions:
(i) $q_{\ell-1} \in \delta(q_{\ell-1},\,\sigma^{s}_\ell)$;
and (ii) $q_{\ell-1} \notin \delta(q_{\ell-1},\,\sigma^{s-}_\ell)$, where~$\sigma^{s-}_\ell =  \sigma^{s-}_\ell \backslash \{s\}$, $\forall s\in \sigma^{s-}_\ell$. 
\hfill $\blacksquare$
\end{definition}

\begin{example}
	\label{example:subtask}
        {As shown in Fig.~\ref{fig:example_decomposable},
          the subtasks associated with~$q_1q_2\\q_4q_5$ are given
          by~$\Omega=\{(1,\texttt{sweep}_{p_{21}} ,\lnot p_{24}),
(1,\texttt{mow}_{p_{21}} ,\\\lnot p_{24}),
(1,\texttt{scan}_{p_{21}}\land\lnot \texttt{mow}_{p_{21}} ,\lnot p_{24})\}$}
\end{example}

In other words, a subtask $(\ell,\,\sigma_\ell,\sigma^s_\ell)$ consists of its index
,its set of position or action propositions and self-loop requirement.
The index should \emph{not} be neglected as the same set of propositions, namely sub-tasks,
can appear multiple times in the run.
It is important to distinguish them by their indices. And the self-loop should be satisfied 
before executing.
Moreover, the two conditions in the above definition require that
each subtask~$\sigma_\ell$ satisfies the segment of the task from $q_\ell$ to $q_{\ell+1}$.
Thus, every element inside~$\sigma_\ell$ needs to be fulfilled for the subtask to be fulfilled.
Note that the decomposition~$\Omega_{\varphi}$ imposes directly a {strict and complete} ordering of the subtasks within,
namely it requires that the subtasks be fulfilled in the exact order of their indices.
This however can be overly restrictive as it prohibits the concurrent execution of several subtasks by multiple agents.
Thus, we propose a new notion of \emph{relaxed and partial} ordering of the decomposition,
as follows.

\begin{definition}[Partial Relations]\label{def:partial}
Given two subtasks in~$\omega_h,
\omega_\ell\in \Omega_{\varphi}$,
the following two types of relations are defined:
\begin{itemize}
\item[(I)] ``less equal'': $\preceq_{\varphi}\subseteq \Omega_{\varphi} \times \Omega_{\varphi}$.
If~$(\omega_h, \omega_\ell)\in \preceq_{\varphi}$ or
equivalently $\omega_h\preceq_{\varphi}\omega_\ell$,
then~$\omega_h$ has to be \emph{started} before $\omega_\ell$ is started.
\item[(II)] ``opposed'': $\neq_{\varphi}\subseteq 2^{\Omega_{\varphi}}$.
If~$\{(\omega_h,\dots,\omega_\ell)\}\subseteq \neq_{\varphi}$
or equivalently $\omega_h\neq_{\varphi}\dots\neq_{\varphi}\omega_\ell$,
then all subtask $\omega_h,\cdots$ cannot all be \emph{executed} simultaneously. 
\hfill $\blacksquare$
\end{itemize}
\end{definition}

\begin{remark}\label{remark:duration}
Note that most related work~\cite{kantaros2020stylus, guo2015multi,
tumova2016multi, luo2021abstraction,luo2021temporal, sahin2019multirobot, jones2019scratchs}
treats the fulfillment of robot actions as \emph{instantaneous},
i.e., the associated proposition becomes True once the action is finished.
Thus, the probability of two actions are fulfilled at the exact same time instant is of measure zero.
The above two relations can be simplified into one ``less than'' relation. 
{The work in~\cite{luo2021temporal} also proposes a partial relation similar to $\preceq_\varphi$ above,
 without considering the opposing relation~$\neq_\varphi$.}
On the contrary, for the action model described in Sec.~\ref{subsec:multi-agent},
each action has a duration that the associated proposition is True during the \emph{whole} period.
Consequently, it is essential to distinguish these two relations defined above,
namely, whether one sub-task should be started or finished before another sub-task. 
\hfill  $\blacksquare$
\end{remark}

\begin{figure}[t!]
\includegraphics[width=0.95\linewidth]{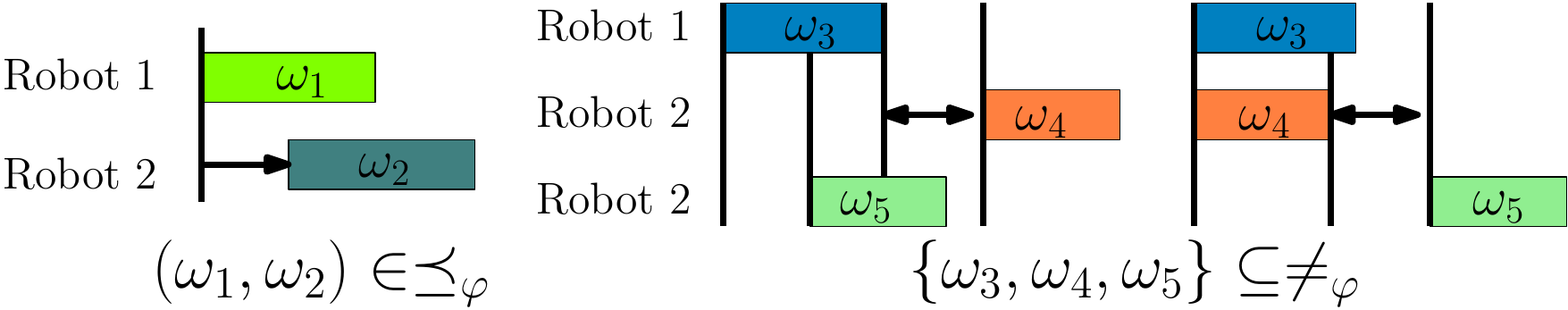}
\centering
\caption{Illustration of the two partial relations
        contained in the poset.
        \textbf{Left}: $\omega_1\preceq_{\varphi} \omega_2$ requires that
task~$\omega_2$ is started after task~$\omega_1$.
\textbf{Middle\&Right}:
$\{\omega_3,\omega_4,\omega_5\}\subset\neq_{\varphi}$
requires that there is no time when
$\omega_3,\omega_4,\omega_5$ are all executing.}
\label{fig:partial}
\end{figure}

The above definition is illustrated in Fig.~\ref{fig:partial}.
Intuitively, the relation~$\preceq_{\varphi}$ represents the ordering
constraints among subtasks,
while the relation~$\neq_{\varphi}$ represents the concurrent constraints.
Given these partial relations above, we can formally introduce the poset of subtasks in~$\Omega_{\varphi}$ as follows.

\begin{definition}[Poset of Subtasks]\label{def:poset}
One partially ordered set (\emph{poset}) over the decomposition $\Omega_{\varphi}$ is given by
\begin{equation}\label{eq:poset}
P_{\varphi} = (\Omega_{\varphi}, \, \preceq_{\varphi}, \, \neq_{\varphi}),
\end{equation}
where~$\preceq_{\varphi}$, $\neq_{\varphi}$ are the partial relations by
Def.~\ref{def:partial}.
\hfill $\blacksquare$
\end{definition}

Similar to the original notion of poset in~\cite{simovici2008mathematical},
the above relation is irreflexive and asymmetric,
however only partially transitive.
In particular, it is easy to see that
if $\omega_1\preceq_{\varphi} \omega_2$ and $\omega_2\preceq_{\varphi} \omega_3$
hold for $\omega_1,\omega_2,\omega_3\in \Omega_{\varphi}$,
then $\omega_1\preceq_{\varphi}\omega_3$ holds.
However, $\{\omega_1,\omega_2\}\subseteq\neq_{\varphi} $ and $\{\omega_2,\omega_3\}\subseteq\neq_{\varphi}$
can not imply $\{\omega_1,\omega_3\}\subseteq\neq_{\varphi}$.
Due to similar reasons, $\{\omega_1,\omega_2,\\\omega_3\}\in\neq_\varphi$ can not imply $\{\omega_1,\omega_2\}\subseteq\neq_{\varphi}$.
Clearly, given a fixed set of subtasks~$\Omega_{\varphi}$, the more elements
the relations $\preceq_{\varphi}$ and $\neq_{\varphi}$ have,
the more temporal constraints there are during the execution of these subtasks.
This can be explained by two extreme cases:
(i) no partial relations in~$\Omega_{\varphi}$, i.e.,
$\preceq_{\varphi}=\emptyset$ and $\neq_{\varphi}=\emptyset$.
It means that the subtasks in~$\Omega_{\varphi}$ can be executed in any temporal order;
(ii) total relations in~$\Omega_{\varphi}$,
e.g., $\omega_h \preceq_{\varphi} \omega_\ell$
and $\{\omega_h, \omega_\ell \}\subseteq\neq_{\varphi} $, for all $h<\ell$.
It means that each subtask in~$\Omega_{\varphi}$ should only start after
its preceding subtask finishes according to their indices in the original accepting run.
For convenience, we denote by $\preceq_{\varphi} \triangleq \mathbb{F}$
and $\neq_{\varphi} \triangleq 2^\Omega_{\varphi}$ for this case,
where $\mathbb{F}\triangleq \{(i,\, j),\, \forall i,\,j\in [0,\, L] \,\text{and}\; i<j\}$.
As discussed in the sequel,
less temporal constraints implies more concurrent execution of the subtasks,
thus higher efficiency of the overall system.
Thus, it is desirable to find one decomposition and the associated poset
that has few partial relations.

\begin{remark}\label{remark:partial-order-motivation}
The above two relations, i.e., $\preceq_\varphi$ and $\neq_\varphi$,
are chosen in the definition of posets due to following observations:
as illustrated in Fig.~\ref{fig:partial} and explained in Remark~\ref{remark:duration},
these two relations can describe any possible temporal relation
between non-instantaneous subtasks.
More importantly,
they can abstract the key information contained in the structure of NBA.
In particular, given any two transitions in an accepting run,
their temporal constraint can be expressed via the~$\preceq_\varphi$.
Secondly, within each transition, there are often subtasks in the ``negated'' set,
meaning that they can not be executed concurrently,
which can be expressed via the~$\neq_\varphi$ relation.
\hfill $\blacksquare$
\end{remark}

\begin{remark}\label{remark:compare-poset}
It is worth noting that the proposed notion of posets contains
the ``decomposable states'' proposed in~\cite{schillinger2018simultaneous}
as a \emph{special case}.
More specifically, the set of decomposable states divide an accepting run into
fully independent segments, where
(i) any two alphabets within the same segment are fully ordered;
(ii) any two alphabets within different segments are not ordered thus independent.
In contrast, the proposed poset allows also independent alphabets within the
same segment.
This subtle difference leads to more current executions not only by
different segments but also within each segment,
thus increases the overall efficiency. \hfill $\blacksquare$
\end{remark}

To satisfy a given poset, the language of the underlying system is
much more restricted.
In particular, the language of a poset is defined as follows.

\begin{definition}[Language of Poset]\label{def:language-poset}
Given a poset $P_\varphi=(\Omega_{\varphi}, \, \preceq_{\varphi}, \, \neq_{\varphi})$,
its language is defined as the set of all finite words
that can be generated by the subtasks in~$\Omega_{\varphi}$
while satisfying the partial constraints.
More concretely, the language is given by
$L(P_\varphi)=\{W_{\varphi}\}$, where $W_{\varphi}$ is a finite
\emph{word} constructed with the set of subtasks in $P_\varphi$, i.e.,
\begin{equation}\label{eq:poset-language}
W_{\varphi}=(t_1,\omega_1) (t_2,\omega_2)\cdots (t_L,\omega_L),
\end{equation}
where the subtask~$\omega_\ell \in \Omega_\varphi$ and
$t_\ell$ is the starting time of subtask $\omega_\ell$.
Furthermore, $W_{\varphi}$ should satisfy the partial relations in $P_\varphi$, namely:
(i) $t_\ell \leq t_{\ell'}$ holds,
$\forall (\omega_\ell,\,\omega_{\ell'})\in \preceq_{\varphi}$;
(ii) $\forall \{\omega_{i}\}\subseteq \neq_{\varphi},
\exists\omega_{\ell},\omega_{\ell'}\in\{\omega_{i}\}, t_\ell + d_\ell <  t_{\ell'}$ holds,
where $d_\ell$ and $d_{\ell'}$ are the durations of
subtasks~$\omega_\ell, \omega_{\ell'} \in \Omega_{\varphi}$.
With a slight abuse of notation, $W_\varphi$ can also denote the
simple sequence of alphabets~$\omega_1\omega_2\cdots \omega_L$.
\hfill $\blacksquare$
\end{definition}

Given the above definition, a poset $P_\varphi$ is called \emph{accepting}
if its language satisfies the original task specification, i.e.,
$L(P_\varphi)\subseteq L_\varphi$.
In other words, instead of directly searching for the accepting word of~$\varphi$,
we can focus on finding the accepting poset that
requires the least completion time.
In the rest of this section, we present how this can be achieved efficiently
in real-time.
First of all, it is worth pointing out that it is computationally expensive to
generate the \emph{complete} set of all accepting poset
and then select the optimal one.
More precisely, even to generate \emph{all} accepting runs in $\mathcal{B}_{\varphi}^-$,
the worst-case computational complexity is~$\mathcal{O}(|Q^-|!)$.
Instead, we propose an anytime algorithm that can generate at least one
valid poset within any given time budget,
while incrementally adding more posets as time allows.

\begin{algorithm}[h]\footnotesize
	\caption{$\texttt{compute\_poset}(\cdot)$: Anytime algorithm to compute accepting posets.}
	\label{alg:compute-poset}
	\SetKwInOut{Input}{Input}
	\SetKwInOut{Output}{Output}
	\Input {Pruned NBA~$\mathcal{B}_{\varphi}^{-}$, time budget $t_0$.}
	\Output{Posets $\mathcal{P}_{\varphi}$, language~$\mathcal{L}_{\varphi}$.}
	Choose initial and final states $q_0\in Q_0^-$ and $q_f\in Q_F^-$;\\
	Set $\mathcal{L}_{\varphi}=\mathcal{P}_{\varphi}=\emptyset$;\\
	Begin modified DFS to find an accepting run $\rho$;\\
	\While{$time<t_0$}{
		\tcc{Subtask decomp. by Def.~\ref{def:subtasks}}
		Compute~ $\Omega$ and word~$W$ given~$\rho$;
		\label{alg-line:decompose}\\
		\If{$W \notin \mathcal{L}_\varphi$\label{alg1:inlp}}{
			Set~$P=(\Omega,\, \preceq_{\varphi}=\mathbb{F},\neq_{\varphi}=2^{\Omega})$,
			and $\preceq_{\varphi}=\mathbb{F}$;\label{alg-line:initial}\\
			Set~$L(P)=Que=\{W\}$ and~$I_1=I_2=\emptyset$;\\
			\tcc{Reduce partial relations}
			\While{$|Que|>0$\label{alg1:begin_que}}{
				$W \leftarrow Que.pop()$;\\
				\For{$i=1,2,\cdots,|W|-1$}{
					$\omega_{1}=W[i]$, $\omega_{2}=W[i+1]$;\\
					$W' \leftarrow $ Switching~$\omega_{1}$
					and~$\omega_{2}$ within~$W$\label{alg1:switch};\\
					\eIf{$W'$ is accepting\label{alg1:waccepting}}{
						
						Add $(\omega_{1},\,\omega_{2})$ to $I_1$;\\
						Add $W'$ to $Que$ if not in $Que$;\\
						Add $W'$ to $L(P)$ if not in $L(P)$;\label{alg1:add-w}	 
					}{Add $(\omega_{1},\,\omega_{2})$ to $I_2$;\label{alg1:end_que}}
				}
				Remove $\{I_1\,\backslash\, I_2\}$
				from~$\preceq_{\varphi} $;\label{alg-line:remove}}
				\For{$\{\omega_i\} \subseteq \neq_{\varphi}$\label{alg-line:calculate_neq}}{
				$W' \leftarrow $ Replace~$\omega_i\in\{\omega_i\}$ in~$W$
				by~$\bigcup\omega_i$;\label{alg1:word-neq}\\
				\If{$W'$ is accepting\label{alg1:waccepting-neq}}{
					Remove~$\{\omega_i\}$
					from~$\neq_{\varphi}$;\label{alg-line:remove_neq}\\}
			}
			\tcc{Self-loop calculation }\For{$W$ in $L(P)$}{
				Get path $\rho'$ by $\sigma_\ell$ of $W'$;\\
				\For{$i=0,1,\cdots,|\rho'|$\label{alg-line:self-loop}}{
					$\sigma^p_{\ell_1}=\sigma^p_{\ell_1}\cap\delta^{-1}(\rho'[i-1],\rho'[i-1])$\\
			}}
			Add~$P$ to~$\mathcal{P}_{\varphi}$,
			add~$L(P)$ to~$\mathcal{L}_{\varphi}$;}
		Continue the modified DFS, and update~$\rho$;
	}
	\Return{$\mathcal{P}_{\varphi}$,~$\mathcal{L}_{\varphi}$};
\end{algorithm}

As summarized in Alg.~\ref{alg:compute-poset},
the proposed algorithm builds upon the modified depth first search (DFS)
algorithm with local visited sets~\cite{sedgewick2001algorithms}.
Given the pruned automaton~$\mathcal{B}_{\varphi}^-$,
the modified DFS can generate an accepting run $\rho$ given the chosen pair of
initial and final states.
Given $\rho$, the associated set of subtasks $\Omega$ and word $W$ can be
derived by following Definition~\ref{def:subtasks}, see Line~\ref{alg-line:decompose}.
Then, a poset $P$ is initialized as $P=(\Omega,\, \mathbb{F},\, 2^{\Omega})$
in Line~\ref{alg-line:initial},
namely, a fully-ordered poset as described after Definition~\ref{def:poset}.
Furthermore, to reduce the partial relations,
we introduce a ``swapping'' operation to change the order of adjacent alphabets,
and then check if the resulting new word can lead to an accepting run.
If so, it means the relative ordering of this two adjacent subtasks can
potentially be relaxed or removed from~$\preceq_{\varphi}$ as in Line~\ref{alg-line:remove}.
On the contrary, for any other word within~$L(P)$, if such swapping does not
result in an accepting run, it is definitively kept in the partial ordering.
The resulting poset is a new and valid poset that have less partial
ordering constraints.
Furthermore, for any subtasks set that belong to the ``opposed'' relation,
a new word is generated by allowing all subtasks to be fulfilled simultaneously
in Line~\ref{alg1:word-neq}. 
If this new word is accepting, it means that this subtasks set do not belong to the
relation~$\neq_{\varphi}$ in Line~\ref{alg-line:remove_neq}. After that, self-loop $\sigma^s_\ell$ is calculated
by checking all feasible word in line~\ref{alg-line:self-loop}.
Note that the resulting~$P$ is only one of the posets and the associated language is
given by~$L(P)$ as defined in~Definition~\ref{def:language-poset}.
Lastly, as time allows, the DFS continues until a new accepting run is found,
which is used to compute new posets with same steps.

\begin{example}
	Continuing from example~\ref{example:subtask}, we can get a poset with 
	$\preceq_\varphi=\{(1,2),(1,3)\},\neq_{\varphi}=\{\{2,3\}\}$.
\end{example}

\begin{lemma}\label{lemma:accepting-poset}
Any poset within $\mathcal{P}_{\varphi}$ obtained by Alg.~\ref{alg:compute-poset}
is accepting.
\end{lemma}
\begin{proof}
Due to the definition of accepting poset, it suffices to show that the
language~$L(P)$ derived above for reach~$P$ is accepting.
To begin with, as shown in Line~\ref{alg1:add-w}, any word~$W$ added to~$L(P)$
is accepting.
Second, assume that there exists a word~$W\in L_\varphi$ but $W\notin L(P)$, i.e.,
$W$ satisfies the partial ordering constraints in~$P$ but does not belong to $L(P)$.
Regarding the ordering relation~$\preceq_{\varphi}$,
due to the iteration process of~$Que$ in Line~\ref{alg1:begin_que}-\ref{alg1:end_que},
any accepting word~$W$ that satisfies the ordering constraints will be added
to~$L(P)$.
In other words, starting from the initial word~$W_0$ associated with the run~$\rho$,
there always exists a sequence of switching operation in Line~\ref{alg1:switch} that results
in the new word~$W$.
With respect to each ordering relation within~$\neq_{\varphi}$, it is even simpler
as any word within~$L(P)$ satisfies this relation and is verified to be accepting
after augmenting the alphabets with the union of all alphabets.
Thus, if~$W\in L_{\varphi}$, it will be first added to~$Que$ in
Line~\ref{alg1:begin_que}-\ref{alg1:end_que} and then
verified in Line~\ref{alg1:word-neq}, thus~$W\in L(P)$. This completes the proof.
\end{proof}

Since Alg.~\ref{alg:compute-poset} is an anytime algorithm,
its output~$\mathcal{P}_{\varphi}$ within the given time budget
could be much smaller than the actual complete set of accepting posets.
Consequently, if a word~$W$ does not satisfy
any poset~$P\in \mathcal{P}_{\varphi}$, i.e.,~$w\notin \mathcal{W}_{\varphi}$,
it can still be accepting.
Nonetheless, it is shown in the sequel for completeness analyses that
given enough time, Alg.~\ref{alg:compute-poset} can generate
the complete set of posets.
In that case, any word that does not satisfy any poset
within~$\mathcal{P}_{\varphi}$ is surely not accepting.
It means that the complete set of posets~$\mathcal{P}_{\varphi}$ is
equally expressive as the original NBA $\mathcal{B}^-$.
The above analysis is summarized in lemma~\ref{lemma:complete-poset}.

\begin{lemma}\label{lemma:complete-poset}
The outputs of Alg.~\ref{alg:compute-poset}
satisfy that~$L(P_i)\cap L(P_j)=\emptyset$, $\forall i\neq j$,
and $L(P_i)\subset  \subset L_\varphi$, $\forall P_i \in \mathcal{P}_\varphi$.
Moreover, given enough time~$t_0\rightarrow \infty$,
the complete set of posets can be returned, i.e.,
$L_\varphi = \cup_{P_i\in \mathcal{P}_\varphi}\,L(P_i)=L_\varphi$.
\end{lemma}
\begin{proof}
The first part can be proven by contradiction.
 Assume that there exists two posets~$P_1,\,P_2\in \mathcal{P}_\varphi$
 and one accepting word~$W\in L_\varphi$ such that~$w\in L(P_1) \cap \in L(P_2)$ holds.
Since the set of subtasks within the poset is simply
 the union of all subtasks within each word,
 it implies that~$\Omega_1 = \Omega_2$.
 Then, as discussed in Lemma~\ref{lemma:accepting-poset},~$W\in L(P_1)$ implies
 that there exists a sequence of switching operations that maps the original
 word~$W_0$ to~$W$, all of which satisfy the partial relations in~$P_1$.
 The same applies to~$W\in L(P_2)$.
 Since the set of subtasks are identical, it implies that the relations in~$P_1$
 is a subset of those in~$P_2$, or vice versa.
 However, since the~$Que$ in Alg.~\ref{alg:compute-poset} iterates through all
 accepting words of the same~$\Omega$,
 the partial relations are the maximum given the same~$\Omega$,
 Thus both partial relations in~$P_1, P_2$ can only be equal and~$P_1=P_2$ holds.
Regarding the second part,
the underlying DFS search scheme in Alg.~\ref{alg:compute-poset} is guaranteed
 to exhaustively find all accepting runs of~$\mathcal{B}^-_{\varphi}$.
 In other words, the complete set of posets~$\mathcal{P}_{\varphi}$ returned
 by the algorithm after full termination is ensured to cover all
 accepting words of the underlying NBA.
 As discussed earlier, the pruning procedure does not effect
 the complete set of accepting words, given the model of the multi-agent system.
 Thus, it can be concluded that the returned language set~$L_\varphi$
 is equivalent to the original task specification.
\end{proof}

\begin{figure}[t!]
	\includegraphics[width=0.9\linewidth]{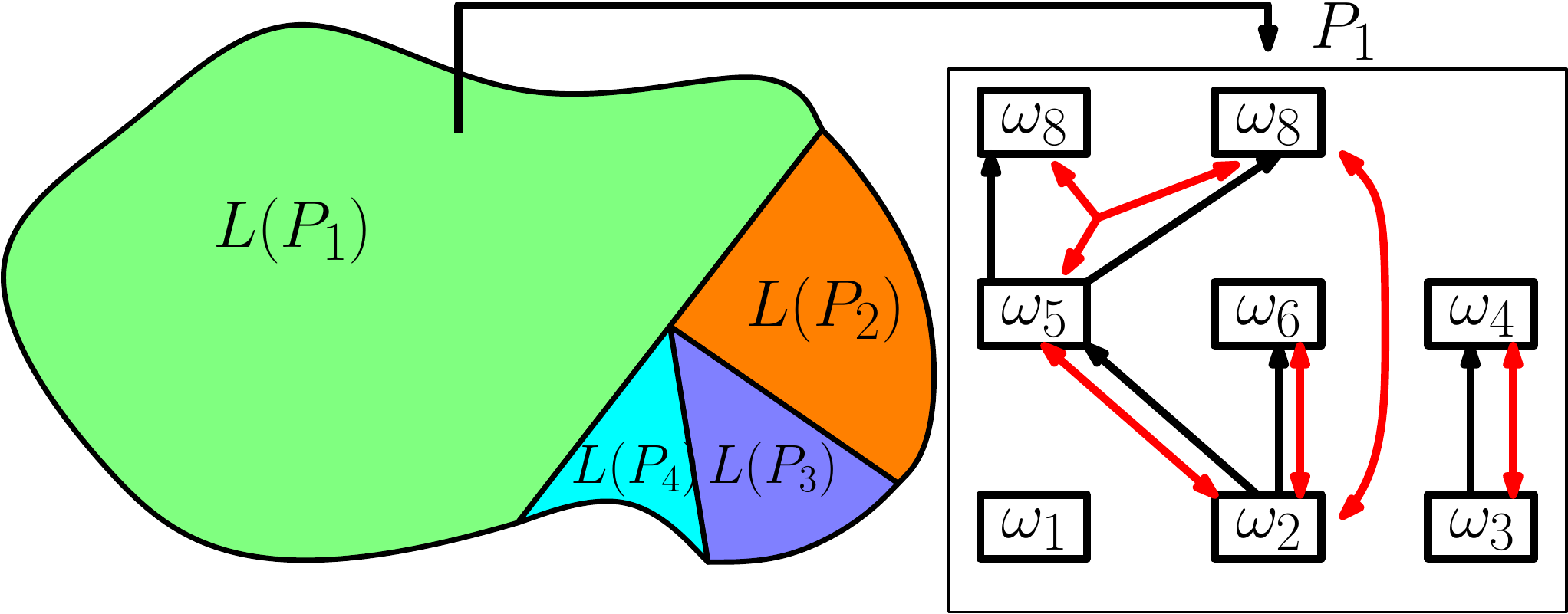}
	\centering
\caption{\textbf{Left}:
an illustration of the relation between the accepting
language of different posets~$L(P_i)$
and the accepting language of the task~$L_\varphi$.
\textbf{Right}:
an example of the poset graph~$\mathcal{G}_{P_\varphi}$,
where the relations~$\preceq_{\varphi}$ and~$\neq_{\varphi}$
are marked by black and red arrows, respectively.}
\label{fig:poset_language}
\end{figure}

Last but not least, similar to the Hasse diagram in~\cite{simovici2008mathematical},
the following graph can be constructed given one poset~$P_{\varphi}$.

\begin{definition}[Poset Graph]\label{def:poset-graph}
The poset graph of $P_{\varphi}=(\Omega,\,\preceq_{\varphi},\,\neq_{\varphi},
\,\Omega_0)$ is a digraph $\mathcal{G}_{P_\varphi}=(\Omega,\,E,\,R)$,
where $\Omega$ is the set of nodes;
$E\subset \Omega \times \Omega$ is the set of directed edges;
$R\subset 2^\Omega$ is the set of undirected special 'edges' which 
connect multiple nodes instead of only two.
A edge $(\omega_1,\,\omega_2)\in E$ if two conditions hold:
(i) $(\omega_1,\, \omega_2)\in \preceq_{\varphi}$;
{and} (ii) there are no intermediate nodes~$\omega_3$ such that
$\omega_1\preceq_{\varphi} \omega_3 \preceq_{\varphi} \omega_2$ holds;
lastly, $\Omega_0\subseteq \Omega$ is set of root nodes that do not have
incoming edges. An undirected 'edge' $(\omega_1,\omega_2,\dots)\in R$, if $\{\omega_1,\omega_2,\dots\}\in \neq_{\varphi}$.
\hfill $\blacksquare$
 \end{definition}

The poset graph~$\mathcal{G}_{P_\varphi}$ provides a straightforward
representation of the partial ordering among subtasks,
i.e., from low to high in the direction of edges.
Note that~$\mathcal{G}_{P_\varphi}$ can be dis-connected with multiple root nodes.
An example of a poset graph is shown in Fig.~\ref{fig:poset_language}.


\subsection{Task Assignment}\label{subsubsec:task-assignment}
Given the set of posets~$\mathcal{P}_{\varphi}$ derived from the previous
section, this section describes how this set can be used to compute
the optimal assignment of these subtasks. More specifically, we consider
the following sub-problems of task assignment:

\begin{problem}\label{problem:}
Given any poset $P=(\Omega,\, \preceq_{\varphi},\, \neq_{\varphi})$
where~$P\in \mathcal{P}_{\varphi}$,
find the optimal assignment of all subtasks in~$\Omega$ to the multi-agent system
$\mathcal{N}$ such that
(i) all partial ordering requirements in $\preceq_{\varphi},\, \neq_{\varphi}$ are
respected; (ii) the maximum completion time of all subtasks is minimized.
\hfill $\blacksquare$
\end{problem}

To begin with, even without the requirements of partial ordering
and collaborative actions, the above problem includes the multi-vehicle routing
problem~\cite{gini2017multi, khamis2015multi},
and the job-shop scheduling problem~\cite{brucker1994branch} as special instances.
Both problems are well-known to be NP-hard.
Thus, the above problem is also NP-hard and its most common and straightforward solution
is to formulate a Mixed Integer Linear Program (MILP). 
{However, there are two major drawbacks of MILP:
(i) the computation complexity grows exponentially with the problem size;
(ii) there is often no intermediate solution before the optimal solution is generated via a MILP solver,
, e.g., CPLEX~\cite{lima2010ibm}.
Both drawbacks hinder the usage of this approach in large-scale real-time applications,
where a timely good solution is far more valuable than the optimal solution.}

\begin{figure}[t!]
	\centering
	\includegraphics[width=0.9\linewidth]{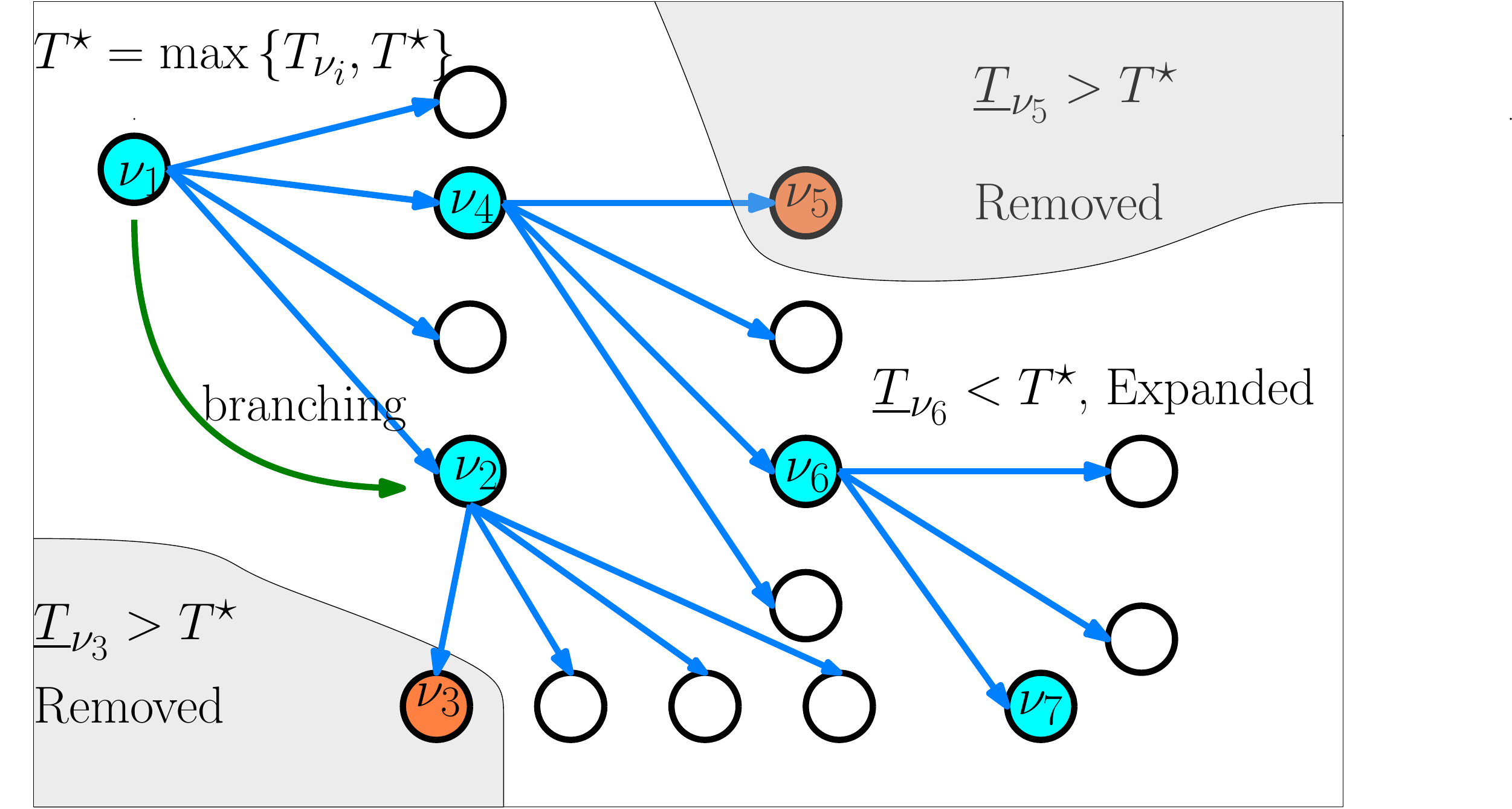}
	\caption{
		Illustration of the main components in the BnB search,
		i.e., the node expansion and branching to generate explore new nodes (in green arrow);
		and the lower and upper bounding to avoid undesired branches (in orange).}
	\label{fig:bnb_search_logic}
\end{figure}

Motivated by these observations, an anytime assignment
algorithm is proposed in this work based on the Branch and Bound (BnB) search
method~\cite{lawler1966branch, morrison2016branch}.
It is not only complete and optimal, but also anytime, meaning that a good
solution can be inquired within any given time budget.
As shown in Fig.~\ref{fig:bnb_search_logic},
the four typical components of a BnB
algorithm are the node expansion, the branching method,
and the design of the upper and lower bounds.
These components for our application are described in detail below.

\textbf{Node expansion}.
Each node in the search tree stands for one partial assignment of the
subtasks, i.e.,
\begin{equation}\label{eq:node}
\nu = (\tau_1,\,\tau_2,\cdots,\tau_N),
\end{equation}
where $\tau_n$ is the ordered sequence of tasks assigned to agent~$n\in \mathcal{N}$.
To give an example, for a system of three agents,
$\nu=((\omega_1,\omega_2),(),())$ means that two subtasks
$\omega_1,\, \omega_2$ are assigned to agent~$1$,
whereas no tasks to agents~$2$ and $3$.

To avoid producing infeasible nodes, we assign only the next task satisfies
the partial order. Let~$\nu$ be the current node of the search tree.
The \emph{next} subtask~$\omega$ to be assigned is chosen from the
poset~$\mathcal{G}_{P_\varphi}$ defined in Def.~\ref{def:poset-graph}
if \emph{all} of its parent subtasks are already assigned, i.e.,
\begin{equation}\label{eq:next-task}
\omega' \in \Omega_\nu, \; \forall \omega' \in \text{Pre}(\omega),
\end{equation}
where~$\text{Pre}(\omega)$ is the set of preceding or parenting subtasks of $\omega$ in~$\mathcal{G}_{P_\varphi}$;
and $\Omega_\nu$ is the set of assigned subtasks as
\begin{equation}\label{eq:node-tasks}
\Omega_{\nu}=\{\omega\in\tau_n,\,\forall n\in \mathcal{N}\},
\;\Omega^-_{\nu} = \Omega\backslash \Omega_{\nu},
\end{equation}
where~$\Omega$ is the set of subtasks from $P_\varphi$;
$\Omega_{\nu}$ is the set of subtasks already assigned in node~$\nu$;
and $\Omega^-_{\nu}$ are the remaining unassigned subtasks.Once this 
subtask~$\omega$ is chosen, the succeeding or child node $\nu^+$
of $\nu$ in the search tree is created by assigning local action to any agent or 
cooperative task combination of agents which is with capable functions. 

\textbf{Branching}.
Given the set of nodes to be expanded,
the branching method determines the order in which these child nodes are
visited.
Many search methods such as breadth first search (BFS), depth first search (DFS)
or $A^\star$ search can be used.
We propose to use $A^\star$ search here as the heuristic function matches well
with the lower bounds introduced in the sequel.
More specifically, the set of child nodes is expanded in the order of
estimated completion time of the whole plan given its current assignment.

\begin{algorithm}[t]
\caption{$\texttt{upper\_bound}(\cdot)$: Compute the upper bound of solutions
rooted from a node}
\label{alg:upper_bound}
\SetKwInOut{Input}{Input}\SetKwInOut{Output}{Output}
\Input {Poset~$P_{\varphi}$, node~$\nu$.}
\Output {Assignment~$\overline{J}_\nu$, upper bound~$\overline{T}_\nu$.}
\While(\tcp*[f]{\eqref{eq:node-tasks}}){$\Omega^-_\nu \neq \emptyset$}{
\ForAll{$\omega\in \Omega^-_\nu$\, \text{and}\, $\{n\}\subset\mathcal{N} $ satisfy $\omega$
\label{algline:all-next}}
{Compute and save $\nu_{n,\omega}$ by assigning $\omega$ to agent set~$\{n\}$ }
Select~$\nu^{+}_{\star}=\textbf{argmax}_{\nu\in \{\nu_{n,\omega}^+\}}\,\{\eta_{\nu}\}$
\label{algline:compute-eta}\tcp*{\eqref{eq:node-makespan}}
$\nu \leftarrow \nu^{+}_{\star}$; \label{algline:expand-node}\\
}
Compute and makespan~$T_\nu$ and subtask start time $T_\Omega$ of assignment~$J_\nu$ \\
\label{algline:return}
\textbf{Return} $\overline{J}_\nu=J_\nu,\, \overline{T}_\nu=T_\nu$,$\overline{T}_\Omega=T_\Omega$;\\
\end{algorithm}

\textbf{Lower and upper bounding}.
The lower bound method is designed to check whether a node fetched by \emph{branching}
has the potential to produce a better solution.
And the upper bound method is tried for each chosen node to update 
 the current best solution.
More specifically, given a node~$\nu$,
the upper bound of all solutions rooted from this node
is estimated via a greedy task assignment policy,
as summarized in Alg.~\ref{alg:upper_bound}:
\begin{equation}\label{eq:upper-bound}
\overline{J}_\nu,\, \overline{T}_\nu ,\overline{T}_\Omega= \texttt{upper\_bound}(\nu,\, P_{\varphi}),
\end{equation}
{where~$\overline{T}_\nu$ is upper bound, and~$\overline{J}_\nu$ is the
associated complete assignment with the same structure of $\nu$, while its $\Omega^-$ is empty; $\overline{T}_\Omega$ is the beginning time
of each subtasks.}
From node~$\nu$, any task~$\omega\in \Omega^-_\nu$ is assigned
to any allowed agent set~$\{n\}\subset\mathcal{N}$ in Line~\ref{algline:all-next},
thus generating a set of child nodes~$\{\nu^+_{n,\omega}\}$.
Then, for each node~$\nu\in \{\nu^+_{n,\omega}\}$,
its \emph{concurrency} level~$\eta_{\nu}$ is estimated as follows:
\begin{equation}\label{eq:node-makespan}
T_\nu = \max_{n\in\mathcal{N}} \{T_{\tau_n}\},\;
T^{\texttt{s}}_\nu = \sum_{\omega\in\Omega_\nu}\, D_{\omega}N_\omega,\;
\eta_\nu = \frac{T^{\texttt{s}}_\nu}{T_\nu},
\end{equation}
where node~$\nu=(\tau_1,\cdots,\tau_N)$;~$T_{\tau_n}$ is the execution
time of all subtasks in~$\tau_n$ by agent~$n$;
$T_\nu$ is the max current makespan calculate by \eqref{eq:node-makespan};
and $T^{\texttt{s}}_\nu$ is the total execution time of all subtasks~$\omega$
given its duration~$D_\omega$ and the number of participants~$N_\omega$.
Thus, the child node with the highest~$\eta_{\nu}$ is chosen as
the next node to expand in Line~\ref{algline:compute-eta}-\ref{algline:expand-node}.
This procedure is repeated until no subtasks remain unassigned.
Afterwards, once a complete assignment~$J_\nu$ is generated, its makespan $T_v$ and start time $T_\Omega$ are obtained by
solving a simple linear program as in Line~\ref{algline:return}.

Furthermore, the lower bound of the makespan of all solutions rooted from this
node is estimated via two separate relaxations of the original problem:
one is to consider only the partial ordering constraints while ignoring
the agent capacities;
another is vice versa. The details of optimization functions for the upper(lower) bound 
can be found in the supplementary material.

\begin{algorithm}[t]
\caption{$\texttt{BnB}(\cdot)$: Anytime BnB algorithm for task assignment}
\label{alg:BnB}
\SetKwInOut{Input}{Input}\SetKwInOut{Output}{Output}
\Input {Agents $\mathcal{N}$, poset $P_{\varphi}$, time budget $t_0$.}
\Output {Best assignment $J^\star$ and makespan $T^\star$.}
Initialize root node~$\nu_0$ and queue $Q=\{(\nu_0,0)\}$; \label{algline:init-start}\\
Set $T^\star=\infty$ and $J^\star=()$;\label{algline:init-end}\\
\While{($Q$ not empty) and ($time<t_0$)}{
Take node $\nu$ off $Q$;\\
$\overline{J}_\nu,\, \overline{T}_\nu,\overline{T}_\Omega  = \texttt{upper\_bound}(\nu,\, P_{\varphi})$
\tcp*{Alg.~\ref{alg:upper_bound}} \label{algline:upper}

\If{$T^\star > \overline{T}_\nu$ \label{algline:update1}}{
Set $T^\star=\overline{T}_\nu$ and $J^\star=\overline{J}_\nu$, $T^\star_\Omega=\overline{T}_\Omega$;\label{algline:update2}}
Expand child nodes $\{\nu^+\}$ from $\nu$ ;\\
\ForAll{$\nu^+\in\{\nu^+\}$}{
	$\underline{T}_\nu = \texttt{lower\_bound}(\nu^+,\,P_{\varphi})$
 \label{algline:lower}\\
	\If{$\underline{T}_\nu \leq T^\star$ \label{algline:store1}}{
			Add $(\nu^+,\underline{T}_\nu)$ into $Q$.\label{algline:store2}
	}
	}
}
\textbf{Return} $J^\star, T^\star, T^\star_\Omega$;
\end{algorithm}


\begin{remark}\label{remark:none-milp}
The computation of both the upper and lower bounds are designed
to be free from any integer optimization.
This is intentional to avoid unpredictable solution time caused by
external solvers.
\hfill $\blacksquare$
\end{remark}

Given the above components, the complete BnB algorithm can be stated as
in Alg.~\ref{alg:BnB}.
In the initialization step in
Line~\ref{algline:init-start}-\ref{algline:init-end},
the root node~$\nu_0$ is created as an empty assignment,
the estimated optimal cost~$T^{\star}$ is set to infinity,
and the queue to store un-visited nodes $Q$ and the lower bound as indexes contains only $(\nu_0,0)$.
Then, within the time budget, a node $\nu$ is taken from $Q$ for expansion with smallest lower bound.
We calculate the upper bound of $\nu$ in line~\ref{algline:upper} and update the optimal value $T^\star,J^\star,T^\star_\Omega$
in line~\ref{algline:update1}-\ref{algline:update2}.
After that, we expand the child nodes $\{\nu^+\}$ of current node $\nu$.
Finally, we calculate the lower bound of each new node $\nu^+$ in line~\ref{algline:lower} and only store
the node with potential to get a better solution into $Q$ in line~\ref{algline:store1},\ref{algline:store2}.
This process repeat until time elapsed or the whole search tree is
exhausted.

\begin{lemma}\label{lemma:BnB-satisfying}
Any task assignment~$J^\star$ obtained from Alg.~\ref{alg:BnB} satisfies
the partial ordering constraints in~$P_{\varphi}$.
\end{lemma}
\begin{proof}
Since the assignment~$J^\star$ belongs to the set of solutions obtained from
the upper bound estimation in Alg.~\ref{alg:upper_bound}
at certain node in the search tree, it suffices to show that any solution
of Alg.~\ref{alg:upper_bound} satisfies the partial ordering constraints.
Regardless of the current node~$\nu$, the set of remaining subtasks
in~$\Omega^-_\nu$ is assigned strictly following the preceding order
in the poset graph as defined in~\eqref{eq:next-task}.
In other words, for any pair~$(\omega_1,\,\omega_2)\in \preceq_{\varphi}\cap \neq_{\varphi}$,
if $\omega_1\in \Omega_\nu$ and $\omega_2\in \Omega^-_\nu$,
then the starting time of~$\omega_2$ is larger than the finishing time of~$\omega_1$.
Similar arguments hold for~$\preceq_{\varphi}$ and~$\neq_{\varphi}$ separately.
\end{proof}

\subsection{Overall Algorithm}\label{subsubsec:overall-algorithm}

\begin{algorithm}[t]
	\caption{Complete algorithm for time minimization
        under collaborative temporal tasks}
	\label{alg:complete}
	\SetKwInOut{Input}{Input}
        \SetKwInOut{Output}{Output}
	\Input {Task formula~$\varphi$, time budget~$t_0$.}
	\Output {Assignment~$J^\star$,  makespan~$T^\star$}
	Compute $\mathcal{B}_{\varphi}$ given~$\varphi$ \\
	Compute $\mathcal{B}^{-}_{\varphi}$ by pruning~$\mathcal{B}_{\varphi}$
        \tcp*{Sec.~\ref{subsubsec:NBA-pruning}}
        Initialize $\mathcal{J}=\emptyset$;\\
        \While{$time<t_0$}{
	$P_{\varphi}\leftarrow \texttt{compute\_poset}(\mathcal{B}^{-}_{\varphi})$
        \label{algline:comp-poset} \tcp*{Alg.~\ref{alg:compute-poset}}
        $(J,\, T, \,T_\Omega)\leftarrow \texttt{BnB}(P_{\varphi})$
        \label{algline:BnB} \tcp*{Alg.~\ref{alg:BnB}}
        Store $(J,\,T,\,T_\Omega)$ in $\mathcal{J}$;\\
        }
        Select~$J^\star$, $T^\star_\Omega$ with minimum $T^\star$ among $\mathcal{J}$;\\
        Get time list $T_\Omega$ for each task;\\
	\textbf{Return} $J^\star$, $T^\star$, $T^\star_\Omega$;
\end{algorithm}

The complete algorithm can be obtained by combining
Alg.~\ref{alg:compute-poset} to compute posets
and Alg.~\ref{alg:BnB} to assign sub tasks in the posets.
More specifically, as summarized in Alg.~\ref{alg:complete},
the NBA associated with the given task~$\varphi$ is derived and pruned as
described in Sec.~\ref{alg:compute-poset}.
Afterwards, within the allowed time budget~$t_0$,
once the set of posets~$\mathcal{P}_{\varphi}$ derived from
Alg.~\ref{alg:compute-poset}
is nonempty, any poset $P_\varphi\in \mathcal{P}_{\varphi}$ is fed to
the task assignment Alg.~\ref{alg:BnB} to compute the current best
assignment~$J$ and its makespan~$T$, which is stored in
a solution set~$\mathcal{J}$.
This procedure is repeated until the computation time elapsed.
By then, the optimal assignment~$J^\star$ and its makespan~$T^\star$
are returned as the optimal solution.

\begin{remark}\label{remark:parallel}
It is worth noting that even though Alg.~\ref{alg:compute-poset} and
Alg.~\ref{alg:BnB} are presented sequentially in
Line~\ref{algline:comp-poset}~-~\ref{algline:BnB}.
They can be implemented and run {in parallel}, i.e., more posets are
generated and stored in Line~\ref{algline:comp-poset},
while other posets are used for task assignment in Line~\ref{algline:BnB}.
Moreover, it should be emphasized that Alg.~\ref{alg:complete} is an \emph{anytime}
algorithm meaning that it can run for any given time budget and generate the
current best solution.
As more time is allowed, either better solutions are found or confirmations are
given that no better solutions exist.
\hfill  $\blacksquare$
\end{remark}

Finally, once the optimal plan~$J^\star$ is computed with the format
defined in~\eqref{eq:node}, i.e.,
the action sequence $\tau_n$ and is assigned
to agent~$n$ with the associated time stamps~$t_{\omega_1}t_{\omega_2}\cdots t_{\omega_n}$ in $T^\star_\Omega$.
In other words,
agent~$n$ can simply execute this sequence of subtasks at the designated time,
namely $\omega_k$ at time~$T_{\omega_k}$.
Then, it is ensured that all task can be fulfilled in minimum time.
However, such way of execution can be prone to uncertainties in system model
such as fluctuations in action duration
and failures during execution, which will be discussed in the next section.

\section{Online Adaptation}
\label{subsec:online-adaptation}

Since there are often uncertainties in the model, e.g., the agents may
transit faster or slower due to disturbances due to the additional constrains of self-loop $\sigma^s_\ell$,
an action might be finished earlier or latter,
or failures may occur during mission,
the optimal plan derived above might be invalid during online execution.
Thus, in this section, we first analyze these uncertainties in the execution time
and agent failures, for which online adaptation methods are proposed.

\subsection{Online Synchronization under Uncertain Execution Time}\label{subsubsec:uncertain}
Uncertainty in the execution time can cause delay or early termination of subtasks.
Without proper synchronization, the consequences can be disastrous.
For instance, one collaborative action is started without waiting for one
delayed collaborator, or one subtask~$\omega_2$ is started before another
subtask~$\omega_1$ is finished, which violates the partial ordering
constraints~$\omega_1\neq_{\varphi} \omega_2$.
These cases would all lead to a failed task execution.
To overcome these potential drawbacks, we propose an adaptation algorithm
that relies on \emph{online synchronization} and distributed communication.

More specifically, consider the optimal assignment~$J^\star$ and the local
sequence of subtasks for agent~$n$: $\tau_n=\omega^1_n\omega^2_n\cdots \omega^{K_n}_n$.
Without loss of generality, agent~$n$ just finished executing~$\omega^{k-1}_n$ and
is during the transition to perform subtask~$\omega^k_n$ at the designated region.
No matter how much the transition is delayed or accelerated,
the following synchronization procedure can be enforced to
ensure a correct execution of the derived plan even under uncertainties:

(i) \emph{Before} execution.
In order to start executing~$\omega^k_n$,
a ``\texttt{start}'' synchronization message is sent by agent~$n$ to agent~$m$,
for each subtask~$\omega^\ell_m$
satisfying~$(\omega^k_n,\,\omega^\ell_m)\in \preceq_{\varphi}$.
This message indicates that the execution of~$\omega^k_n$ is started
thus~$\omega^\ell_m$ can be started.
On the other hand, for each subtask~$\omega^\ell_m$
satisfying~$(\omega^\ell_m, \omega^k_n)\in \preceq_{\varphi}$,
agent~$n$ waits for the ``\texttt{start}'' synchronization message
from agent~$m$.
This message indicates that the execution of~$\omega^\ell_m$ is started
thus~$\omega^k_n$ can be started.
Last, for each subtask~$\omega^\ell_h$
satisfying~$\{\omega^k_n,\omega^\ell_h,\dots\}\subseteq \neq_{\varphi}$,
agent~$n$ checks the ''\texttt{start}'' or ``\texttt{stop}'' synchronization message
from agent~$h$ to ensure that not all subtasks in $\{\omega^k_n,\omega^\ell_h,\dots\}$ is executing.
Moreover, the self-loop constrain associated with the subsequent subtasks must be
followed to satisfy the poset.
Namely, each subset of agents that are executing~$\omega_{\ell'}$ publishes
the self-loop requirement~$\sigma^s_\ell$ associated with the subsequent subtask $\omega_\ell$.
It means that all other agents would satisfy the requirements in~$\sigma^s_\ell$
before~$\omega_{\ell'}$ is finished.

\begin{figure}[t]
	\centering
	\includegraphics[width=0.8\linewidth]{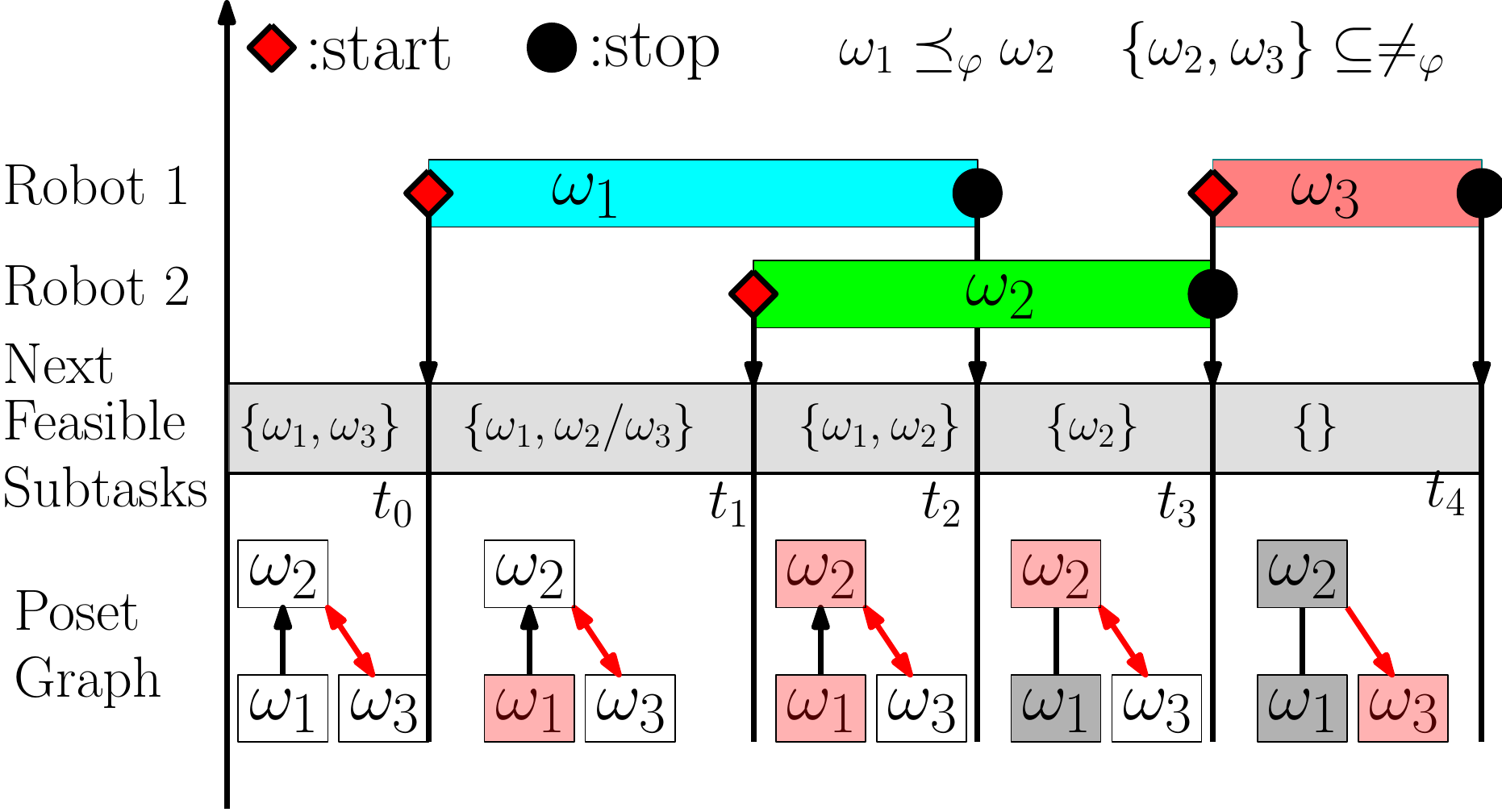}
        \caption{
Illustration of the online synchronization process
described in Sec.~\ref{subsubsec:uncertain}.
Consider the constraints~$\omega_1 \preceq_{\varphi}\omega_2$
and~$\{\omega_1, \omega_3\}\subseteq\neq_\varphi $.
The ``\texttt{Start}'' and ``\texttt{Stop}'' messages are
marked by red diamonds and black circles, respectively. The agent can only
execute the subtask if all relevant partial relation is satisfied.}
\label{fig:online adaption}
\end{figure}

(ii) \emph{During} execution.
If~$\omega^k_n$ is a collaborative action, then
agent~$n$ sends another synchronization message to each collaborative agent
to start executing this action.
Otherwise, if~$\omega^k_n$ is a local action, then agent~$n$ starts the
execution directly;
(iii) \emph{After} execution.
After the execution of~$\omega^k_n$ is finished,
for each subtask~$\omega^\ell_h$
satisfying~$\{\omega^k_n,\,\omega^\ell_h,\dots\}\subseteq \neq_{\varphi}$,
a ``\texttt{stop}'' synchronization message is sent by agent~$n$ to agent~$h$,
This message indicates that the execution of~$\omega^k_n$ is finished
thus~$\omega^\ell_h$ can be started.
The above procedure is summarized in Fig.~\ref{fig:online adaption}.
The synchronization for collaborative tasks is easier as a collaborative
subtask can start if all participants are present at the location.

\begin{remark}\label{remark:syn}
The synchronization protocol above is event-based,
i.e.,  only the subtasks within the partial relations
are required to synchronize,
which are much less than the complete set of subtasks.
In comparison, the product-based solution~\cite{baier2008principles}
and the sampling-based solution~\cite{kantaros2020stylus}
require full synchronization during task execution,
meaning that the movement of all agents should be synchronized for each
transition in the global plan.
Moreover, the Mixed Integer Linear Program (MILP)-based solution
in~\cite{luo2021temporal, jones2019scratchs}
requires no synchronization as each agent simply executes the subtasks
according to the optimal time plan.
The planning algorithm in~\cite{schillinger2018simultaneous}
also requires no synchronization as the local subtasks of each agent
are designed to be independent (however losing optimality).
As validated in the numerical experiments,
the synchronization protocol offers great flexibility
and robustness against fluctuation in task durations during execution,
while ensuring correctness.
\hfill  $\blacksquare$
\end{remark}

\subsection{Plan Adaptation under Agent Failures}\label{subsubsec:failure}
Whenever an agent is experiencing motor failures that it can not
continue executing its remaining subtasks,
a more complex adaptation method is required as the unfinished subtasks
should be re-assigned to other agents.

Assume that agent~$N_d$ has the optimal
plan~$\tau_{N_d}=\omega^1_{N_d}\omega^2_{N_d}\\\cdots \omega^{K_{N_d}}_{N_d}$.
It fails at time $t=T_d$ after the execution of subtask~$\omega^{k_d-1}_{N_d}$ and during the
transition to subtask~$\omega^{k_d}_{N_d}$.
Consequently, the set of unfinished tasks is given by:
\begin{equation}\label{eq:unfinished}
\widehat{\Omega}_{N_d}=\{\omega^{k'}_{N_d},\,k'=k_d,k_d+1,\cdots,K_{N_d}\},
\end{equation}
which is communicated to other agents before agent~$N_d$ failed.
Note that if an agent fails during the execution of a subtask,
this subtask has to be re-scheduled and thus re-executed.

Given this set of subtasks, the easiest recovery is to recruit another new
agent~$N'_d$ with the same capabilities as agent~$N_d$ and
takes over all tasks in~$\widehat{\Omega}_{N_d}$.
However, this is not always feasible, meaning that~$\widehat{\Omega}_{N_d}$ needs to
be assigned to other existing agents within the team.
Then, the BnB algorithm in Alg.~\ref{alg:BnB} is modified as follows.
First, the initial root node now consists of the subtasks that are already
accomplished by each functional agent, i.e.,
\begin{equation}\label{eq:new-initial}
  \nu'_0=(\tau'_1,\cdots,\tau'_{N_d-1},\tau'_{N_d+1},\cdots,\tau'_{N}),
\end{equation}
where~$\tau'_n=\omega^1_n \omega^2_n\cdots \omega^{K_n}_n$ and $K_n$ is last
subtask be accomplished at time $t=T_d$,
for each agent~$n=1,\cdots,N_d-1,N_d+1,\cdots,N$.
Namely, agent~$N_d$ is excluded from the node definition.
Second, the node expansion now re-assigns all unfinished tasks:
$
  \widehat{\Omega}_d= \bigcup_{n\in \mathcal{N}}\,\widehat{\Omega}_n,
$
where~$\widehat{\Omega}_n\subset \Omega_{\varphi}$ is the set of unfinished tasks
for agent~$n\in \mathcal{N}$, defined similarly as in~\eqref{eq:unfinished}.
Namely, each remaining task is selected according to the \emph{same} partial
ordering constraints~$P_{\varphi}$ as before agent failure, for node expansion.
Afterwards, the same branching rules and more importantly,
the methods for calculating lower and upper bounds are followed.
Consequently, an adapted plan~$\widehat{J}^\star$ can be obtained
from the same anytime BnB algorithm.
It is worth noting that the above adaptation algorithm shares the same
completeness and optimality property as Alg.~\ref{alg:complete}.

Last but not least, when there are multiple failed agents, the above procedure
can be applied with minor modifications, e.g., the node definition excludes
all failed agents.

\section{Algorithmic Summary}\label{subsec:summary}

To summarize, the proposed planning algorithm in Alg.~\ref{alg:complete}
can be used offline to synthesize the complete plan that
minimizes the time for accomplishing the specified collaborative temporal tasks.
During execution, the proposed online synchronization scheme can be applied to
overcome uncertainties in the duration of certain transition or actions.
Moreover, whenever one or several agents have failures, the proposed adaptation
algorithm can be followed to re-assign the remaining unfinished tasks.
In the rest of this section,
we present the analysis of completeness, optimality and computational
complexity for Alg.~\ref{alg:complete}.

\begin{theorem}[Completeness]\label{theo:completeness}
Given enough time, Alg.~\ref{alg:complete} can return the optimal
assignment~$J^\star$ with minimum makespan~$T^\star$.
\end{theorem}
\begin{proof}
 To begin with, Lemma~\ref{lemma:accepting-poset} shows that any poset obtained
 by Alg.~\ref{alg:compute-poset} is accepting.
 As proven in Lemma~\ref{lemma:complete-poset},
 the underlying DFS search scheme finds all accepting runs of~$\mathcal{B}^-_{\varphi}$ via exhaustive search.
 Thus, the complete set of posets~$\mathcal{P}_{\varphi}$ returned by
 Alg.~\ref{alg:compute-poset} after full termination is ensured to cover all
 accepting words of the original task.
 Moreover, Lemma~\ref{lemma:BnB-satisfying} shows that any assignment~$J^\star$
 from the BnB Alg.~\ref{alg:BnB} satisfies the input poset
 from Alg.~\ref{alg:compute-poset}.
 Combining these two lemmas, it follows that any assignment~$J^\star$
 from Alg.~\ref{alg:complete} satisfies the original task formula.
 Second, since both Alg.~\ref{alg:compute-poset} and~\ref{alg:BnB} are exhaustive,
 the complete set of posets and the complete search tree of all
 possible assignments under each poset can be visited and thus taken into account
 for the optimal solution.
 As a result, once both sets are enumerated, the derived assignment~$J^\star$ is
 optimal over all possible solutions.
\end{proof}

Second, the computational complexity of Alg.~\ref{alg:complete}
is analyzed as follows. To generate one valid poset in Alg.~\ref{alg:compute-poset},
the worst case time complexity is~$\mathcal{O}(M^2)$, where~$M$ is the maximum
number of subtasks within the given task thus bounded by the number of edges in the
pruned NBA~$\mathcal{B}^-_{\varphi}$.
However, as mentioned in Sec.~\ref{subsec:nba},
the size of~$\mathcal{B}$ is double exponential to the size of~$|\varphi|$.
The number of posets is upper bounded by the number of accepting runs
within~$\mathcal{B}^-_{\varphi}$, thus worst-case combinatorial to
the number of nodes within~$\mathcal{B}^-_{\varphi}$.
Furthermore, regarding the BnB search algorithm, the search space is in the worst
case~$\mathcal{O}(M!\cdot N^M)$ as the possible sequence of all subtasks
is combinatorial and the possible assignment is exponential to the
number of agents.
However, the worst time complexity to compute the upper bound via Alg.~\ref{alg:upper_bound}
remains~$\mathcal{O}(M\cdot N)$ as it greedily assigns the remaining subtasks,
while the complexity to compute the lower bound is~$\mathcal{O}(M^2)$ as it relies on a BFS over
the poset graph~$\mathcal{G}_{P_\varphi}$.
How to decompose the length of an overall formula while ensuring
the satisfaction of each subformula, thus overcoming the bottleneck
in the size of~$\mathcal{B}_{\varphi}$, remains a part of our ongoing work.

\begin{remark}\label{remark:anytime}
The exponential complexity above is expected due to the NP-hardness of the considered problem.
However, as emphasized previously, the main contribution of the proposed algorithm
is the anytime property.
In other words, it can return the best solution within the given time budget,
which is particularly useful for real-time applications where computation time is limited.
\hfill  $\blacksquare$
\end{remark}

\section{Simulations and Experiment}\label{sec:experiments}

This section contains the numerical validation over large-scale multi-agent systems, both in simulation and on actual hardware.
The proposed approach is implemented in Python3 on top of Robot Operating System (ROS) to enable communication across planning, control and perception modules.
All benchmarks are run on a workstation with 12-core Intel Conroe CPU.
More detailed descriptions and experiment videos can be found in the supplementary file.

\subsection{Simulations and Experiment}\label{subsec:simulation}

The numerical study simulates a team of multiple UGVs and UAVs
that are responsible for maintaining a remote photovoltaic (PV) power station.
We first describe the scenario and three types of tasks,
followed by the results obtained via the proposed method.
Then, we introduce various changes in the environment and agent failures,
in order to validate the proposed online adaptation algorithm.
Third, we perform scalability analysis of our method by increasing
the system size and the task complexity.
Lastly, we compare our methods against several strong baselines, in terms of
optimality, computation time and adaptation efficiency.

\begin{figure}
\includegraphics[scale=0.18]{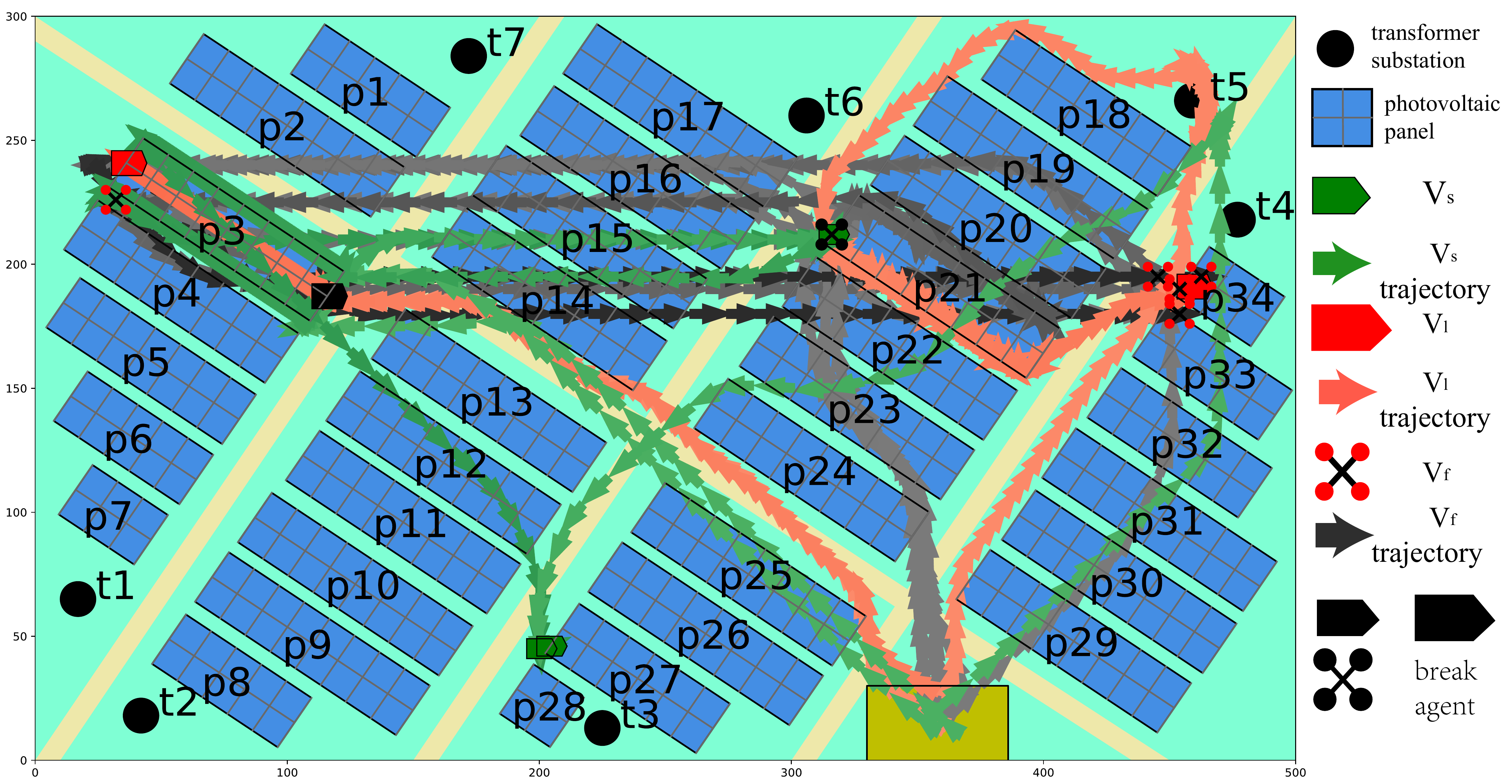}
\caption{Simulated PV power station in the numerical study,
  which consists of PV panels~$\texttt{p}_i$, roads,
  inverters/transformers~$\texttt{t}_i$ and base stations~$b$.
  The arrow trajectories are the path of Agent swarm
  executing the LTL formula~$\varphi_{1}$. And the arrow direction
  is the motion direction and the arrow density correspond to the velocity of different agents.}
\label{fig:workspace}
\end{figure}

\subsubsection{Workspace Description}\label{subsubsec:ws}

Consider a group of UAVs and UGVs that work within a PV power station for
long-term daily maintenance.
As shown in Fig.~\ref{fig:workspace},
the station consists of mainly three parts: PV panels $\texttt{p}_1,\cdots,\texttt{p}_{34}$, roads,
inverter/transformer substations $\texttt{t}_1,\cdots,\texttt{t}_7$ and the robot base station $\texttt{b}$. 
Furthermore, there are one type of UAVs and two types of UGVs as table~\ref{table:agent} showed.
The UAVs $V_f$ are quadcopters which can move freely between all interested places.
The larger type of UGVs, denoted by $V_l$, has the limitation of not going to PV panels or transformers;
the smaller ones $V_s$ can travel more freely, e.g., under the PV panels but not under the transformers. 
As a result, different types of robots have different motion 
model~$\mathcal{G}_n$ as described in Sec.~\ref{subsec:multi-agent}
and distinctive action models.
The traveling time among the regions of interest is estimated by the route
distance and their respective speed. The actions required and the  
Descriptions of these regions of interest and robot actions are summarized in
Table.~\ref{fig:symbols}.
Note that some actions can be performed alone while some require direct
collaboration of several agents,
e.g., one $V_s$ can $sweep$ debris under the PV panel while one $V_l$
and two $V_s$ are required to $repair$ a broken PV panel.

\subsubsection{Task Description}\label{subsubsec:task}

For the nominal scenario, we consider a system of moderate size,
including $12$ agents: 6 $V_f$, 3 $V_l$ and 3 $V_s$.
Scalability analysis to larger systems are performed later in Sec.~\ref{subsubsec:scalable}.
Moreover, we consider a complex task and test it with agent failure.

\begin{table}[t]\footnotesize
\caption{definition of agent with function}
\label{table:agent}
\begin{tabular}{|c|c|c|c|}\hline
	\textbf{Agent type} &\textbf{label} & \textbf{Capable action} & \textbf{Speed}$(m/ s )$\\ \hline
	 Quadcopter& $V_f$  & $temp, scan, wash $ & 10 \\ \hline
	 Larger UGV& $V_l$  & $wash,repair_l,fix$ & 4 \\ \hline
	 Smaller UGV& $V_s$  &  $sweep, mow, repair_s, fix$ & 4 \\ \hline
\end{tabular}
\end{table}

\begin{table}[t]\footnotesize
 \centering
\caption{Description of related regions and agent actions.}
\label{fig:symbols}
\begin{tabular}{|c|m{0.5\columnwidth}|c|}\hline
\textbf{Proposition} & \textbf{Description}\centering & \textbf{Duration} [s]\\ \hline
$\texttt{p}_1,\cdots,\texttt{p}_{34}$ & $34$ PV panels. & $\backslash$ \\ \hline
$\texttt{b}$ & Base stations for all agents to park and charge. & $\backslash$ \\ \hline
$\texttt{t}_1,\cdots,\texttt{t}_7$ & $7$ transformers. & $\backslash$ \\ \hline
$\texttt{temp}_{\texttt{p}_i,\texttt{t}_i}$ &
Measure temperature of panel~$\texttt{p}_i$ and transformer $\texttt{t}_i$.
Requires 1 $temp$ action. & 10 \\ \hline
$\texttt{sweep}_{\texttt{p}_i}$& Sweep debris around any panel~$\texttt{p}_i$.
Requires 1 $sweep$ action. & 190\\ \hline
$\texttt{mow}_{\texttt{p}_i,\texttt{t}_i}$ &
Mow the grass under panel~$\texttt{p}_i$ or transformer~$\texttt{t}_i$.
Requires 1 $mow$ action. & 190\\ \hline
$\texttt{fix}_{\texttt{t}_i}$ &
Fix malfunctional transformer~$\texttt{t}_i$.
Requires 2 $fix$ action collaborations & 72\\ \hline
$\texttt{repair}_{\texttt{p}_i}$ &
Repair broken panel~$\texttt{p}_i$.
Requires 2 $repair_s$,1 $repair_l$ action collaborations . & 576\\ \hline
$\texttt{wash}_{\texttt{p}_i}$ &
Wash the dirt off panel~$\texttt{p}_i$.
Requires 2 $wash$ actions collaborations. & 565\\ \hline
$\texttt{scan}_{\texttt{p}_i,\texttt{t}_i}$ &
Build 3D models of panel~$\texttt{p}_i$ or transformer~$\texttt{t}_i$
for inspection. Requires 3 $scan$ action. & 95\\ \hline
\end{tabular}
\end{table}

This task can be specified as the following LTL formulas, 
requires a series of limited actions to 
maintain the photovoltaic power station:
\begin{equation}\footnotesize
\label{eq:task1}
  \begin{aligned}
\varphi_1 = & \Diamond(\texttt{repair}_{\texttt{p}_{3}} \wedge \lnot \texttt{scan}_{\texttt{p}_3} \wedge\Diamond \texttt{scan}_{\texttt{p}_3})
\wedge \Diamond (\texttt{wash}_{\texttt{p}_{21}} \wedge \\
&\Diamond \texttt{mow}_{\texttt{p}_{21}} \wedge \Diamond \texttt{scan}_{\texttt{p}_{21}}) \wedge \Diamond ( \texttt{sweep}_{\texttt{p}_{21}} \wedge \lnot \texttt{wash}_{\texttt{p}_{21}} \wedge\\
& \Diamond \texttt{mow}_{\texttt{p}_{21}}) \wedge \Diamond(\texttt{fix}_{\texttt{t}_5} \wedge \lnot \texttt{p}_{\texttt{18}}) \wedge \lnot \texttt{p}_{24} \,U \, \texttt{sweep}_{\texttt{p}_{27}} \\
&\wedge \Diamond (\texttt{wash}_{\texttt{p}_{34}} \wedge \bigcirc \texttt{scan}_{\texttt{p}_{34}})
\end{aligned}
\end{equation}

The location of these subtasks are chosen across the workspace
thus a coordination strategy to minimize completion time is crucial for the task.

\begin{figure}[t!] 
		\centering%
		\includegraphics[height = 0.12 \textwidth]{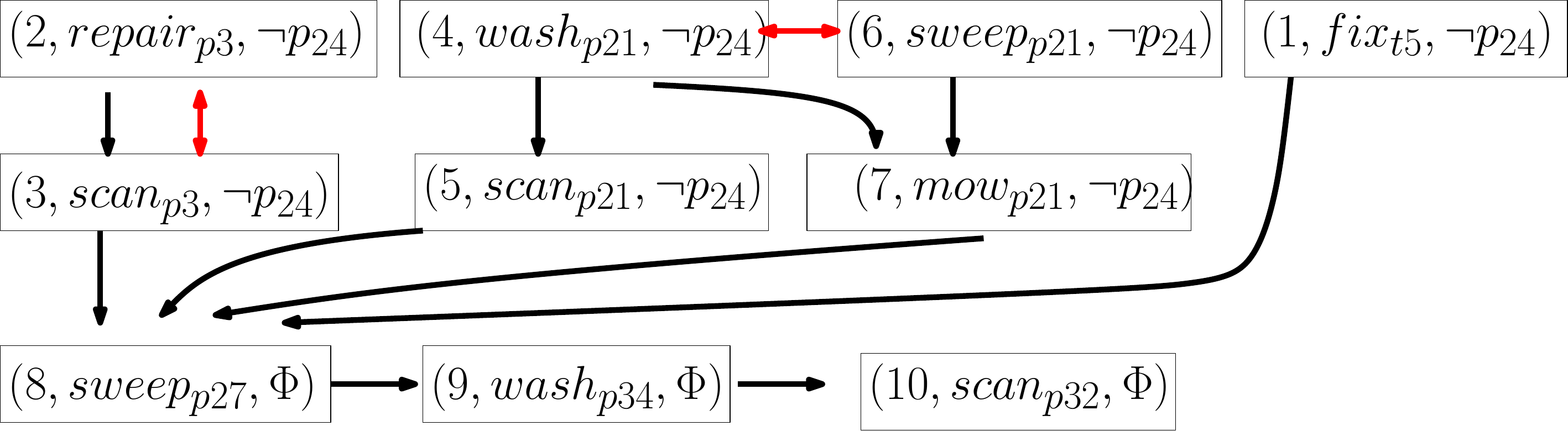}
	\caption{Poset graph of task $\varphi_1$, in which the negative labels of $\sigma_\ell$ are omitted here for simply.
          The relations~$\preceq_\varphi,\, \neq_{\varphi}$ are marked
          by black and red arrows, respectively.}
       \label{fig:task12-posets}
\end{figure}

\subsubsection{Results}\label{subsubsec:results}
In this section, we present the results of the proposed method for
the above tasks, including the computation of posets,
task assignment via the BnB search algorithm, and the task execution results.

\textbf{Partial analysis}:
The NBA~$\mathcal{B}_{\varphi_1}$ associated with task one in~\eqref{eq:task1}
contains~$707$ states and~$16044$ edges. And the pruning step reduces $84.9\%$ edges within 
$30.43$ second. Then, the Alg.~\ref{alg:compute-poset} explores 
 $4$ accepting runs in $0.14s$ to find the first poset and gets the one of the best poset in $22.40s$.
 Finally showed in \ref{fig:task12-posets}, we choose the best poset $P^{p}_{\varphi}$ with $10$ subtasks, whose language 
 $L(P^{p}_{\varphi})$ has $525$ \emph{Words}. In $P^{p}_{\varphi}$, there are 
 multiple subtasks can be executed in parallel such as $\omega_2 $ with $ \omega_4$,$\omega_1 $ with $ \omega_7$.
 However, these subtasks are still ordered as no subtask set can be executed independent with the left 
 subtasks. That means we cannot
 divide the word into series independent parts with the method in \cite{schillinger2018simultaneous}.
 It's worth noting that this poset has only one subtask $\omega_7$ with $\texttt{mow}_{\texttt{p}_{21}}$,
 which is required in twice in $\varphi_1$. And it follows additions partial orders as
 $\omega_4\preceq_\varphi \omega_6,\omega_2\preceq_\varphi \omega_6$. That means our method found 
 a more efficient poset with relations not explicitly written in the formula.
 There are two $\neq_\varphi$ relations as $(\omega_1,\omega_3),(\omega_2,\omega_4)$, due to the constrains 
 $\texttt{repair}_{\texttt{p}_3}\wedge\lnot\texttt{scan}_{\texttt{p}_3} $ and
  $\texttt{sweep}_{\texttt{p}_{21}} \wedge \lnot \texttt{wash}_{\texttt{p}_{21}}$ in formula $\varphi_1$.
  Before execute $\omega_8$, all subtasks has the self-loop constrains $\lnot p_{24}$ due to
   $\lnot \texttt{p} U \texttt{sweep}_{\texttt{p}_{27}}$.

\begin{figure}[t!]
\centering%
\includegraphics[width = 0.50\textwidth]{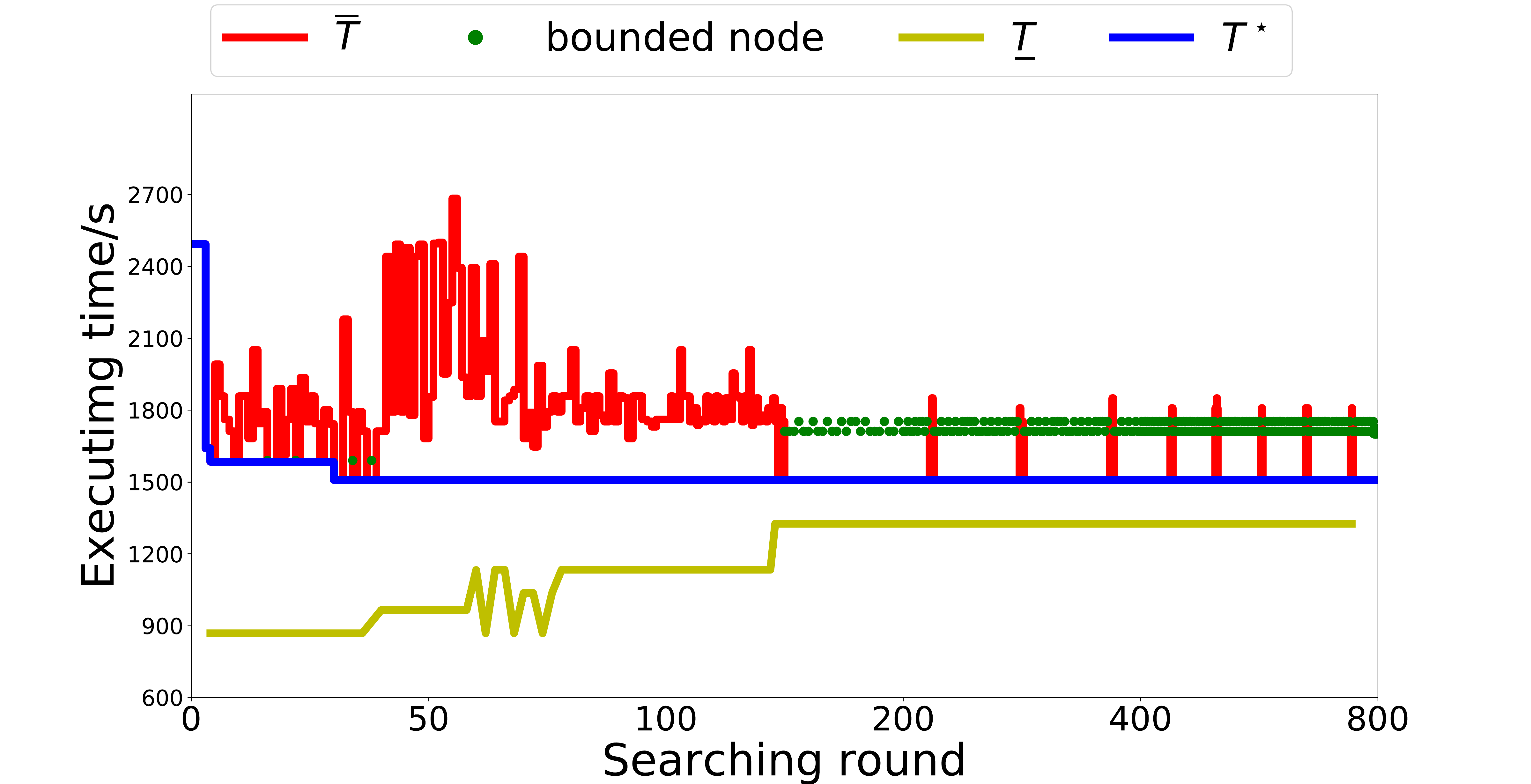}
\caption{Illustration of the upper and lower
bounds~$\overline{T}_\nu,\,\underline{T}_\nu$, and the optimal value~$T^\star$,
along with the BnB search process.}
\label{fig:task2-bnb}
\end{figure}
\textbf{Task assignment}
Then, during the task assignment step,
the first valid solution is found in~$0.131s$.
Afterwards, at~$t=1.75s$, a node is reached and its estimated lower
bound is larger than current upper bound, and thus cut off the search tree.
Overall, around~$84.2\%$ of visited nodes are cut off,
which clearly shows the benefits of the ``bounding'' mechanism.
Then, the estimated upper-bound rapidly converged
to the optimal $T^\star=1388.5s$ in~$3.34s$ with exploring~$30$ nodes.
This is due to the branching efficiency during the BnB search,
by using the estimated lower bounds as heuristics.
Lastly, the whole search tree is exhausted after more than~$10$ hours
due to the complexity of the problem.

In the optimal task assignment, for the same type of task $\texttt{wash}$,
different types of agent are employed as $V_{f3},V_{l9}$ for \\$\texttt{wash}_{\texttt{p}_{21}}$ and
 $V_{f4},V_{f5}$ for $\texttt{wash}_{\texttt{p}_{34}}$. And all the constrains 
 of $\preceq_\varphi,\neq_\varphi$ are satisfied such as $\texttt{mow}_{\texttt{p}_{21}}$ 
should be executed after $\texttt{repair}_{\texttt{p}_3}$, $\texttt{sweep}_{\texttt{p}_21}$
should not be execute at same time which denote by triangles of the corresponding color.
What's more, the most subtasks without these relations are executed parallel as $\omega_1,\omega_2$.
These parallelisms dramatically reduce the make span.
In the trajectory graph~\ref{fig:workspace}, the self-loop constrains of each subtasks are all satisfied
as before executing $\omega_8$, no agent is permitted enter the PV panels $P_{24}$, while $V_f, V_s$ are allowed 
crossing other PV panels.

\begin{figure}[t!]
  \begin{minipage}[t]{1\linewidth}
		\includegraphics[height =0.5\textwidth]{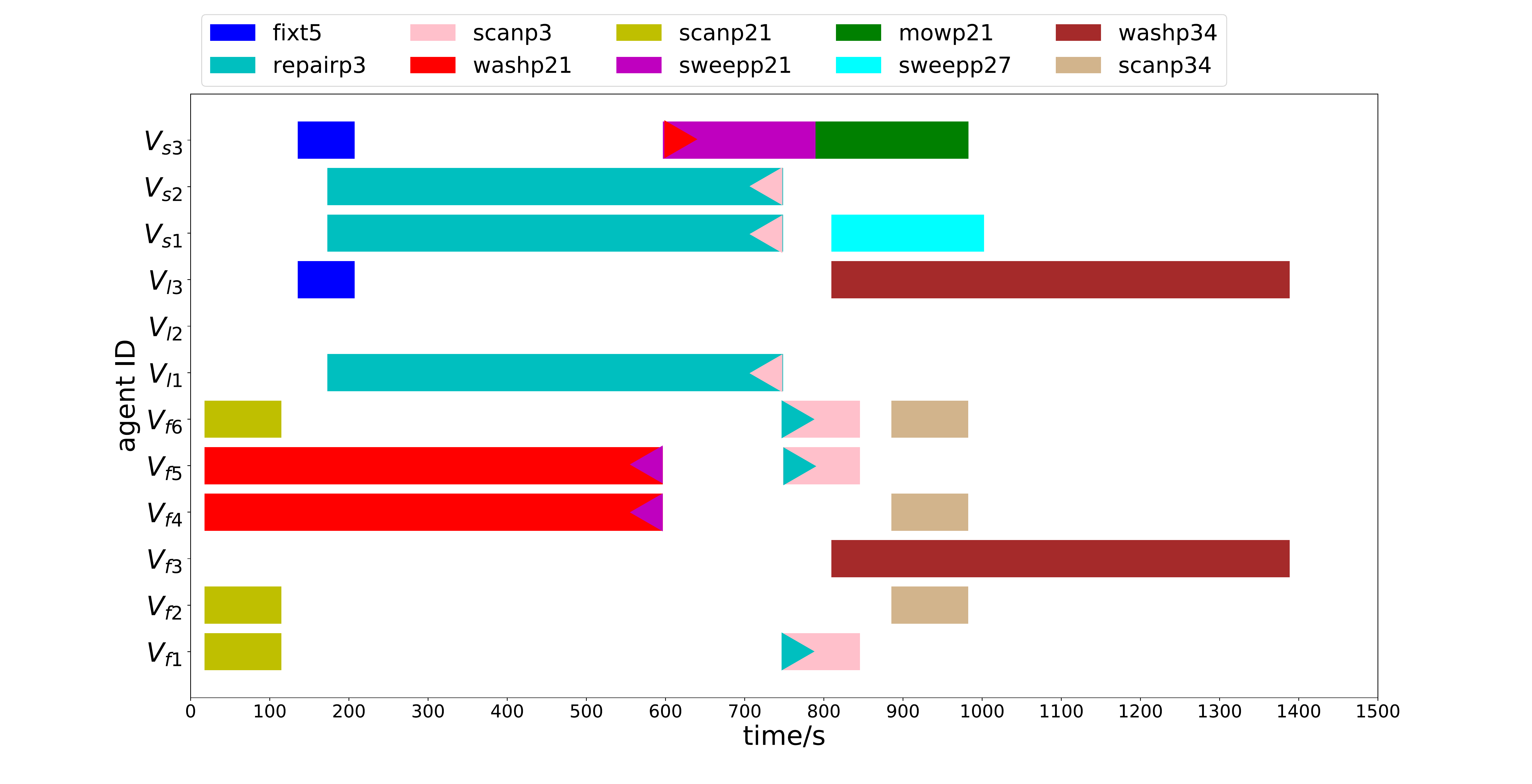}
	
\end{minipage}%
   \centering %
   \label{Gantt_graph}
\caption{ Gantt graph of offline planning. } 
\end{figure}

\subsubsection{Online Adaptation}\label{subsubsec:exp-adapt}
\begin{figure}[t]
	  \begin{minipage}[t]{1\linewidth}
		\includegraphics[height =0.6\textwidth]{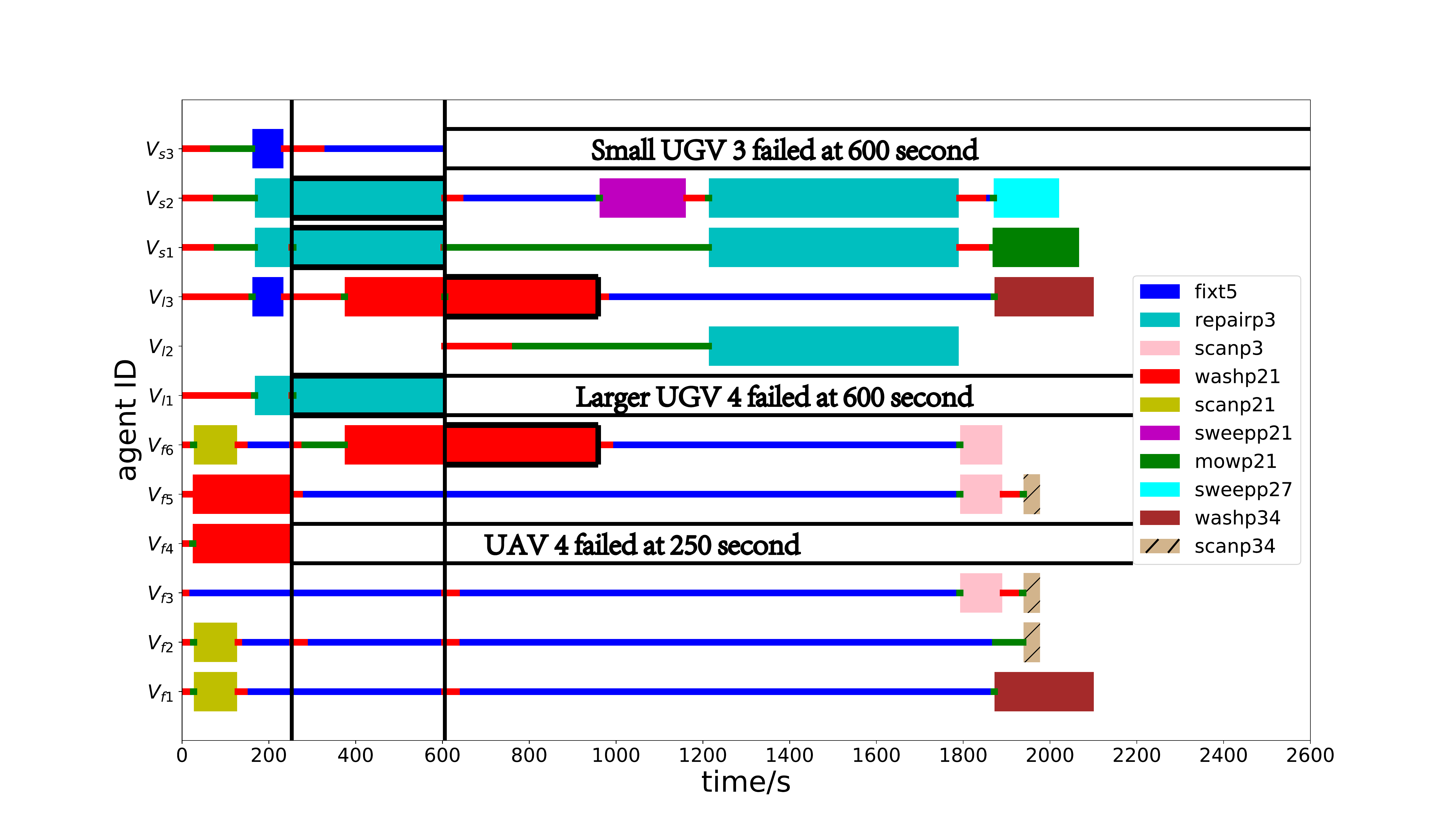}
	\end{minipage}%
	\centering
	\caption{
Gantt graph of the plan execution under agent failures and fluctuated subtask duration during the execution.
Additional lines are added to highlight the online synchronization process. Red seqments indicate that the agents
are during transition among regions, green seqments for "waiting for collaborators", and blue segments for 
“waiting for preceding subtask in partial order”. What's more, three failures agent are highlighted in black.
Interrupted subtasks are repeated in the new-assignment.
    }
        \label{fig:online-failure-task}
\end{figure}

As an important part of the contribution, we now simulate
the following two practical scenarios to validate the proposed online adaptation algorithm.:
(i) fluctuations in the execution time of subtasks;
(ii) several agents break down during the online execution,

First of all, we artificially change the executing time of certain subtasks.
For instance, the executing time of the maintenance tasks for smaller panels
is reduced in comparison to the large panels, e.g., the execution time
of $\texttt{wash}_{\texttt{p}_{34}}$ are reduced to $141s$ from $565s$, as the size
of $\texttt{p}_{34}$ is $25\%$ of~$\texttt{p}_{10}$. And the transfer time 
between different regions will be disturbed.The proposed online synchronization method in
 Sec.~\ref{subsubsec:uncertain} is applied during execution to dynamically accommodate these fluctuations.
As the fig~\ref{fig:online-failure-task} showed, $V_{s_1}, V_{s_2}$ arrived $p_3$ first
and then began "waiting for collaborators" until $V_l1$ arrive $p_3$ and start to executing
  $\texttt{repair}_{\texttt{p}_3}$. And after finished task $\texttt{scan}_{\texttt{p}_{21}}$,
  agent $V_{f_2},V_{f_6}$ go to proper place quickly but cannot start $\texttt{scan}_{\texttt{p}_3}$
  until $\texttt{repair}_{\texttt{p}_3}$ finished. Thus, they turn to the state of "waiting for preceding subtasks
  in partial order".

Secondly, more severe scenarios are simulated where agents break down
during task execution and thus are removed from the team.
More specifically, vehicle~$V_{f_3}$ breaks down at~$200s$, $V_{l_1},\, V_{s_3}$ break
down at~$600s$ during the execution of $\varphi_1$.
Consequently, as shown in Fig.~\ref{fig:online-failure-task}, $\texttt{wash}_{\texttt{p}_{21}}$
is re-assigned to other agent as one of its cooperaters $V_{f_3}$ is failed. And subtask 
$\texttt{repai}$\\$-\texttt{r}_{\texttt{p}_{3}}$ is continuing as its cooperate situation and partial relations 
are still satisfied. As described in Sec.~\ref{subsubsec:failure},
the set of unfinished tasks is re-assigned to the remaining agents by re-identifying
the current node in the BnB search tree and continue the planning process.
It can be seen that no subtasks are assigned to~$V_{f_3}$ anymore in the updated assignment.
Then, as $V_{l_1},\, V_{s_3}$ break down at $600s$, the execution of subtask~$\texttt{repair}_{\texttt{p}_{3}}$ 
is interrupted. And it is executed again by vehicles~$V_{l_2}$ and $V_{s_1},V_{s_3}$ at~$1222s$.
In the same way, the unfinished subtasks are re-assigned to the remaining agents.
It is worth noting that the partial ordering constraints are respected at all
time during the adaptation.
For instance, $V_{f_1}$ cannot execute the subtask~$\texttt{wash}_{\texttt{t}_{34}}$
before~$\texttt{mow}_{\texttt{p}_{21}}$ is started,
as~$\texttt{mow}_{\texttt{p}_{21}}\leq_\varphi\texttt{wash}_{\texttt{p}_{34}}  $ holds.
All subtasks are fulfilled at~$2109s$, despite of the above contingencies.
The trajectory is showed in fig~\ref{fig:workspace}.

\subsubsection{Scalability Analysis}\label{subsubsec:scalable}
To further validate the scalability of the proposed methods,
the following tests are performed:
(i) the same task with increased team sizes,
e.g., $16$, $24$, $32$ and $40$;
(ii) test more LTL formula list with different structure. 

As summarized in Table~\ref{table:more-agents},
as the system size is increased from~$8$ to $40$,
the computation time to obtain the \emph{first}
solution for task Three remains almost unchanged,
while the time taken to compute the optimal value increases slightly.
This result verifies that the proposed anytime algorithm is beneficial
especially for large-scale systems, as it can returns a high quality solution fast,
and close-to-optimal solutions can be returned as time permits.
Secondly, more tasks as $\varphi_2,\varphi_3$ are considered as follows:
\begin{equation}\footnotesize
\label{eq:task2}
\begin{aligned}
\varphi_2 = &\Diamond (\texttt{wash}_{\texttt{p}_{11}} \land \lnot\texttt{scan}_{\texttt{p}_i} \wedge\Diamond \texttt{scan}_{\texttt{p}_{11}}\wedge \Diamond ((\texttt{mow}_{\texttt{p}_{11}}\quad\quad\quad \\
&\wedge\lnot \texttt{wash}_{\texttt{p}_{11}} ) \wedge\Diamond (\texttt{sweep}_{\texttt{p}_{11}} \land \lnot\texttt{mow}_{\texttt{p}_{11}} )))\wedge\\
&\Diamond (\texttt{temp}_{\texttt{p}_{25}} \land \Diamond \texttt{repair}_{\texttt{p}_{25}}\wedge \Diamond ((\texttt{scan}_{\texttt{p}_{25}}\wedge\\
&\lnot \texttt{wash}_{\texttt{p}_{25}} ) \wedge\Diamond (\texttt{sweep}_{\texttt{p}_{25}} \land \lnot{\texttt{p}_{26}} ))) \wedge  \Diamond
\texttt{temp}_{\texttt{t}_4},
\end{aligned}
\end{equation}  

\begin{equation}\footnotesize
\label{eq:task3}
\begin{aligned}
\varphi_3 = &\Diamond (\texttt{temp}_{\texttt{p}_{25}} \land\Diamond \texttt{repair}_{\texttt{p}_{25}}
\wedge \Diamond ((\texttt{scan}_{\texttt{p}_{25}}\wedge
\lnot \texttt{wash}_{\texttt{p}_{25}} ) \\
&\wedge\Diamond (\texttt{sweep}_{\texttt{p}_{25}} \land \lnot{\texttt{p}_{26}} ))) \wedge  \Diamond
\texttt{temp}_{\texttt{t}_4},
\wedge  \Diamond (\texttt{sweep}_{\texttt{p}_{8}}\\& \wedge \Diamond \texttt{wash}_{\texttt{p}_{8}} )\land \Diamond \texttt{repair}_{\texttt{p}_4} \bigcirc \lnot \texttt{p}_4 \wedge \Diamond ( \texttt{sweep}_{\texttt{p}_{8}}\\
& \wedge \lnot \texttt{wash}_{\texttt{p}_{8}} \wedge\Diamond 
\texttt{scan}_{\texttt{p}_{8}} )\wedge \lnot \texttt{temp}_{\texttt{t}_4} \,U \, \texttt{fix}_{\texttt{t}_{4}},
\end{aligned}
\end{equation}
where~$\mathcal{B}_{\varphi_2}$ contains $216$ states
	 and~$\mathcal{B}_{\varphi_3}$ contains~$970$ states.
	As summarized in Table~\ref{table:more-agents} , the computation time of
	 both posets and tasks assignment are increased significantly, as the task becomes more
	 complex. 
	 However, the time when the first solution is obtained in tasks assignment does not monotonically increase due to
	 the polynomial complexity of upper bound method.

\begin{table}[t]\footnotesize
  \centering
  \begin{threeparttable}
	\caption{Scalability analyses of the proposed method}
	\label{table:more-agents}
  \begin{tabular}{|c|c|c|c|}\hline
    \tnote{1} \begin{tabular}{@{}c@{}} System \\ $(V_f,V_s,V_l)$\end{tabular}
    &  \tnote{3} $t_{\varphi_1}\,[s]$
    &  $t_{\varphi_2}\,[s]$
    & \tnote{3} $t_{\varphi_3}\,[s]$  \\[0.5ex] \hline 
		$(8,4,4)$ & $0.13,5.9$ &$0.14,1.08$ & $0.23,4.81$  \\[0.5ex]
		$(12,6,6)$ & $0.15,4.6$  & $0.08,1.85$ & $0.22,5.23$ \\[0.5ex]
		$(16,8,8)$  & $0.13,5.0$  & $0.10,1.56$ & $0.21,4.33$ \\[0.5ex]
		$(20,10,10)$  & $0.53,8.4$ & $0.09,2.11$ & $0.58,6.20$  \\[0.5ex] \hline
		\tnote{2} Poset analysis  & $64.4,71.1$ &$8.4,37.9$ & $136.4,552.5$ \\ [0.5ex]
    \hline
  \end{tabular}
  \begin{tablenotes}
    \item[1] The number of different types of agents as system size increases.
    \item[2] The associated solution time, measured by two time stemps when: the
    first poset is returned, the best poset with largest language is returned.
    \item[3] The associated solution time,
          measured by two time stamps when:
          the first solution is returned,
          and the optimal solution is returned.
  \end{tablenotes}
\end{threeparttable}
\end{table}

\subsubsection{Comparison}\label{subsubsec:compare}
The proposed method is compared against several
state-of-the-art methods in the literature.
More specifically, four methods below are compared:

\textbf{Prod}: the standard solution~\cite{baier2008principles}
that first computes
the Cartesian products of all agent models,
then computes the product B\"uchi automaton,
and searches for the accepting run within.
As the brute-force method,
it is well-known to suffer from complexity explosion.

\textbf{Milp}: the optimization-based solution that
formulates the assignment problem of posets as a MILP
the compute optimal assignment similar
to~\cite{luo2021temporal, jones2019scratchs},
i.e., instead of the search method.
The partial relations are formulated as constraints
in the program.
An open source solver GLPK~\cite{makhorin2008glpk} is used.

\textbf{Samp}: the sampling based method proposed
in~\cite{kantaros2020stylus}.
Compared with the product-based methods, it does not pre-compute
the complete system model. Instead it relies on a sampling strategy
to explore only relevant search space.
However, since it does not support collaborative actions natively,
we modify the definition of transitions there slightly.

\textbf{Decomp}: the task assignment strategy proposed
in~\cite{schillinger2018simultaneous}.
As discussed earlier in Sec.~\ref{sec:introduction},
the proposed task decomposition strategy only allows completely
independent subtasks.
Furthermore, since it does not support collaborative actions,
collaborative subtasks are decomposed manually.
\begin{table}\footnotesize
  \centering
  \begin{threeparttable}
	\caption{Comparison to other methods.}
	\label{table:compare_time}
	\begin{tabular}{|c|c|c|c|c|c|}\hline
	  \tnote{1} Method & \tnote{2} $t_{\texttt{fir}}\, [s]$
          & \tnote{2} $t_{\texttt{opt}}\, [s]$
	  & \tnote{2} $t_{\texttt{fin}}\,[s]$ & \tnote{2}$T_{\texttt{obj}}\,[s]$
          & \tnote{2} $N_{\texttt{sync}}$ \\ \hline
		\multirow{2}{*}{\textbf{Prod}}& $\infty$ & $\infty$ & $\infty$ & -- & -- \\
                 & $\infty$ & $\infty$ & $\infty$ & -- & -- \\
                \hline
		\multirow{2}{*}{\textbf{Milp}} & 2069.27 & 2069.27 & 2069.27 & 1058.47 & -- \\
                &$\infty$ &$\infty$ & $\infty$ & -- & --  \\
                \hline
		\multirow{2}{*}{\textbf{Samp}} & 328.59 & 1838.96 & $\infty$ & 1968.03 & 24 \\
                 & 3280.68 &  16294.30 & $\infty$ & 1968.03 & 24 \\
                \hline
		\multirow{2}{*}{\textbf{Decomp}} & 580.16 & 580.16 & 4581.3 & 1266.99 & 0 \\
		 & 1151.24 & 1151.24 & 5082.07 & 1267.00 & 0 \\
                \hline
		\multirow{2}{*}{\textbf{Ours}} & 24.81 & 25.26 & $\infty$ & 1058.47 & 8 \\
                 & 28.12 & 37.40 & $\infty$ & 1058.47 & 8 \\
		\hline
	\end{tabular}
  \begin{tablenotes}
  \item[1] For each method, the first row  measures the time for the
    system of~$12$ agents while the second row for~$24$ agents.
  \item[2] The time to derive the first solution,
    the time to derive the optimal solution,
    the termination time, the objective as the task completion time,
and the number of synchronizations required.
  \end{tablenotes}
   \end{threeparttable}
\end{table}

Two identical comparisons are performed under different system sizes,
namely $12$ and $24$ agents, to compare not only efficiency but also scalability.
To begin with, the nominal system of~$12$ agents under task~$\varphi_2$ is considered.
The above four methods are used to solve the same planning problem.
As summarized in Table.~\ref{table:compare},
the results are compared in the following four aspects:
the time to derive the first solution,
the time to derive the optimal solution,
the termination time, the optimal solution,
and finally the number of online synchronizations required.
Since the methods \textbf{Prod}, \textbf{Milp} and \textbf{Decomp} are not
anytime, the time to obtain  the first solution equals to the time
when the optimal solution is obtained.
It can be seen that the \textbf{Prod} failed to generate any solution
within~$11h$ as the system-wide product automaton for both cases
has more than $10^{19}$ states.
The \textbf{Milp} method is only applicable for small problems,
which returns the optimal solution in~$0.5h$ but fails to
return any solution within~$16h$ for the large problem.
The \textbf{Samp} method has the anytime property but it takes ten times
the time to generate
the first feasible solution, compared with our method.
In addition, since the subtasks are executed in sequence by the solution,
the actual time of task completion is significantly longer.
The \textbf{Decomp} method can solve both problems but the overall time
for task completion is longer than our results,
which matches our analyses in Remark~\ref{remark:compare-poset}.
In comparison, our method returns the first solution for both cases in less
than~$30s$ and the optimal solution within another~$10s$.
It can be seen that the task completion time remains the same for both cases,
which is consistent with the \textbf{Milp} method.

Last but not least, the last column in Table~\ref{table:compare_time} compares the number
of synchronizations required during execution.
Although the same solution is obtained, the \textbf{Milp} method requires more
synchronization during execution than our method.
This is because our method requires only synchronization for relations within the
posets, rather than all consecutive subtasks.
The \textbf{Prod} and \textbf{Samp} methods require more synchronization due to
their fully sequential execution,
while the \textbf{Decomp} method requires no synchronization as
the local subtasks of each agent are independent.

\subsection{Hardware Experiment}\label{subsec:hardware-experiment}
For further validation with hardware,
a similar set-up as the simulation is built as shown in Fig.~\ref{fig:ws}.
In total~$4$ UAVs and $2$ UGVs are deployed in the workspace of of $4\times5\, m^2$.
Each robot communicates wirelessly to the control PC via ROS,
of which the state is monitored by the OptiTrack system.
Different tasks and scenarios are designed to show
how the proposed methods perform on actual hardware.
Experiment videos can be found in the supplementary file.

\subsubsection{Workspace and Task Description}\label{subsubsec:hw-ws-task}

\begin{figure}[t!]
  \begin{minipage}[t]{0.49\linewidth}
    \centering%
	\includegraphics[width =0.88\textwidth]{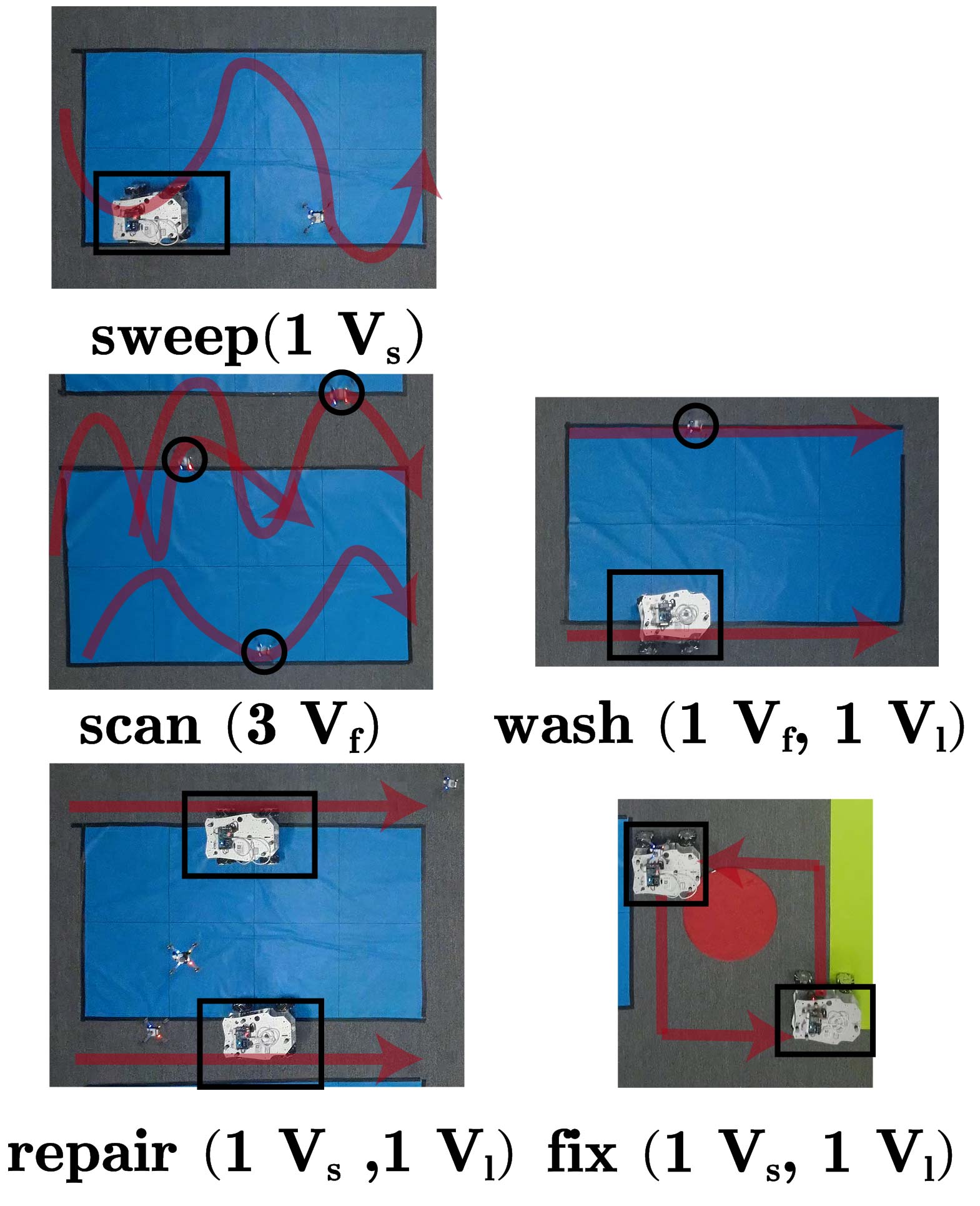}
\end{minipage}%
  \begin{minipage}[t]{0.49\linewidth}
    \centering
	\includegraphics[width =0.9\textwidth]{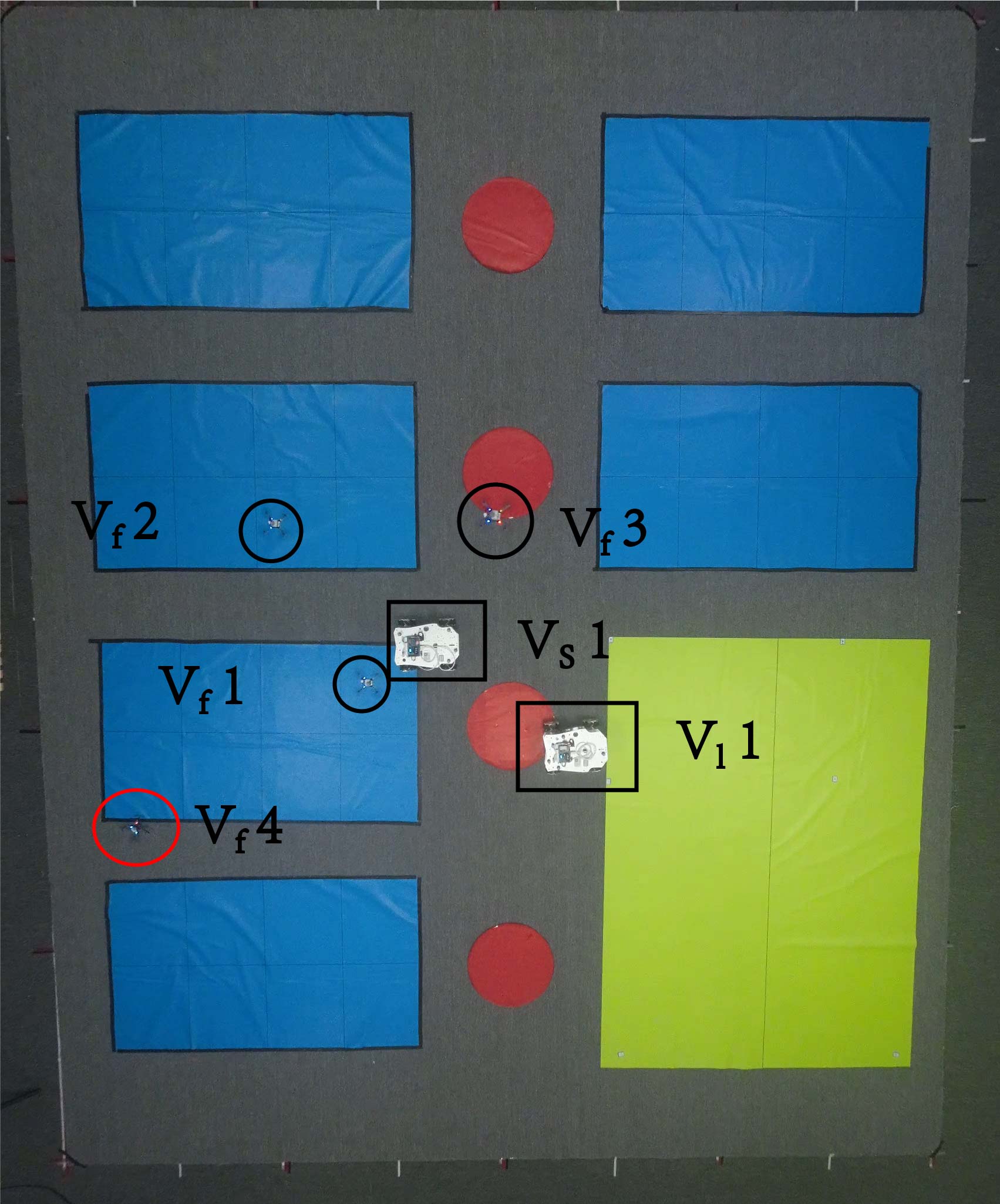}
\end{minipage}%
\caption{Snapshots of the task execution. \textbf{Left}:
Trajectory and collaborators of subtasks.
\textbf{Right}: UAV $4$ (marked in red) is manually stopped to mimic motor failure,
while the rest of the team adapts and continues
the task execution.}
\label{fig:ws}
\end{figure}

The workspace mimics the PV farm described in the numerical simulation.
As showed in Fig.~\ref{fig:ws},
there are~$6$ PV panels ($\texttt{p}_1$-$\texttt{p}_6$, marked in blue),
$4$ transformer substations
($\texttt{t}_1$-$\texttt{t}_4$, marked in red)
and~$1$ base station ($b_1$ marked in yellow).
Moreover,~$4$ UAVs and~$2$ UGVs are deployed to maintain the PV farm,
where the $UAVs$ are Crazyfly mini-drones (denoted by~$V_f$) and
the $UGVs$ are four-wheel driven cars with mecanum wheels (denoted by~$V_s$, $V_l$).
Existing mature navigation controllers are used and omitted here for brevity.
The routine maintenance task considered can be specified with the following LTL formulas:
\begin{equation}\footnotesize
\begin{aligned}
	\varphi_4=& \Diamond(\texttt{repair}_{\texttt{p}_2} \land  \lnot\texttt{scan}_{\texttt{p}_2} \land \Diamond
	\texttt{scan}_{\texttt{p}_2} 
	\land \Diamond (\texttt{sweep}_{\texttt{p}_2} \\ 
	& \land \lnot \texttt{repair}_{\texttt{p}_2}) )\land \Diamond \texttt{fix}_{\texttt{t}_1} \land \Diamond \texttt{scan}_{\texttt{p}_3}
	\land \Diamond \texttt{wash}_{\texttt{p}_5},
\end{aligned}
\end{equation}

\subsubsection{Results}\label{subsubsec:hw-results}
\begin{figure}[t!]
	\centering
	\includegraphics[width=0.45\textwidth]{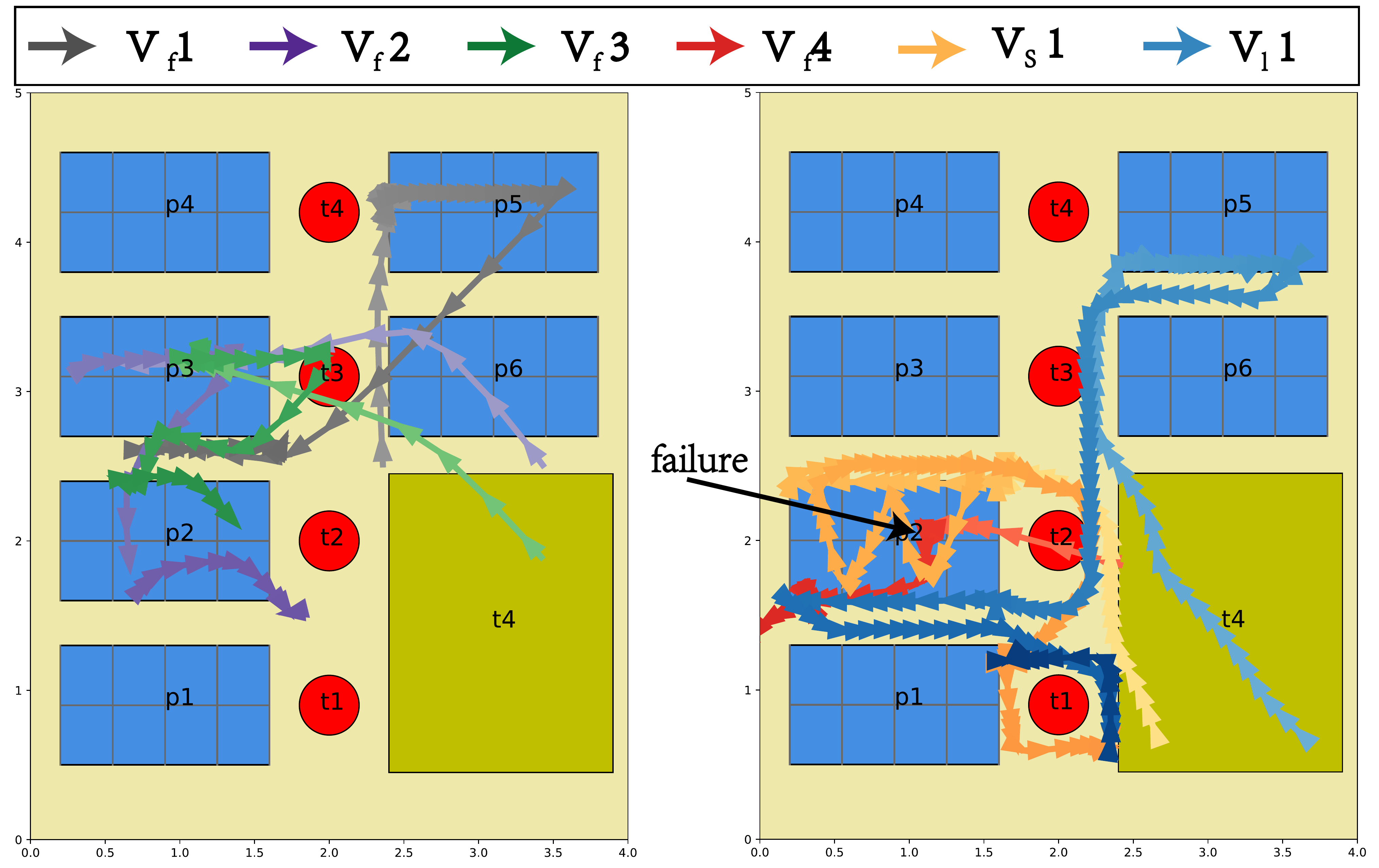}
	\caption{Agent trajectories during the task execution.
          \textbf{Left}: The normal scenario.
          \textbf{Right}: UAV~$4$ is manually taken down at~$75s$.}
        \label{fig:exp-trajs}
\end{figure}

First, we describe the nominal scenario.
Following the procedure described in Alg.~\ref{alg:complete},
the LTL formula is converted to its NBA~$\mathcal{B}$ with $62$ nodes and $521$ edges.
And the pruned NBA~$\mathcal{B}^-$ has $62$ nodes and $377$ edges.
Only one poset is found with Alg.~\ref{alg:compute-poset},
which contains~$6$ subtasks whose language~$L(P)$ equals to the full language~$L(\mathcal{B}^-)$.
Furthermore, Alg.~\ref{alg:upper_bound} finds the
optimal task assignment within~$3.7s$,
which has the estimated makespan of~$124s$, after exploring~$59$ nodes.
During execution, it is worth noting that due to collision avoidance and communication delay,
the fluctuation in the time of navigation and task execution is \emph{significant}.
Consequently, the proposed online synchronization protocol in Sec.~\ref{subsubsec:uncertain}
plays an important role to ensure that the partial constraints
are respected during execution,
instead of simply following the optimal schedule.
The execution of the complete task lasts~$170s$ and
the resulting trajectories are shown in Fig.~\ref{fig:exp-trajs}.
Moreover, to test the online adaptation procedure
as described in Sec.~\ref{subsubsec:failure},
one UAV~$V_{f_4}$ is stopped manually to mimic a motor failure during execution at~$75s$,
as shown in Fig.~\ref{fig:ws}.
During adaptation, a new node is located the BnB search tree
given the set of unfinished tasks
and the search is continued until a new plan is found within~$0.8s$.
As a result, UAV~$V_{f_1}$ takes over the subtask~$\texttt{scan}_{\texttt{P}_2}$
to continue the overall mission.
The resulting trajectory is shown in Fig.~\ref{fig:exp-trajs},
where the trajectory of~$V_{f_4}$ before failure is shown in red,
and the trajectory in blue is another UAV taking over the subtasks.
In the end, the complete task is accomplished in $178s$.

\section{Conclusion}\label{sec:conclusion}

In this work, a novel anytime planning algorithm has been proposed for
the minimum-time task planning of multi-agent systems
under complex and collaborative temporal tasks.
Furthermore, an online adaptation algorithm has been proposed to tackle
fluctuations in the task duration and agent failures during online execution.
Its efficiency, optimality and adaptability have been validated extensively via
simulations and experiments.
Future work includes the distributed variant.

\bibliographystyle{IEEEtran}
\bibliography{contents/references}

\begin{thebibliography}{10}
\providecommand{\url}[1]{#1}
\csname url@samestyle\endcsname
\providecommand{\newblock}{\relax}
\providecommand{\bibinfo}[2]{#2}
\providecommand{\BIBentrySTDinterwordspacing}{\spaceskip=0pt\relax}
\providecommand{\BIBentryALTinterwordstretchfactor}{4}
\providecommand{\BIBentryALTinterwordspacing}{\spaceskip=\fontdimen2\font plus
\BIBentryALTinterwordstretchfactor\fontdimen3\font minus
  \fontdimen4\font\relax}
\providecommand{\BIBforeignlanguage}[2]{{%
\expandafter\ifx\csname l@#1\endcsname\relax
\typeout{** WARNING: IEEEtran.bst: No hyphenation pattern has been}%
\typeout{** loaded for the language `#1'. Using the pattern for}%
\typeout{** the default language instead.}%
\else
\language=\csname l@#1\endcsname
\fi
#2}}
\providecommand{\BIBdecl}{\relax}
\BIBdecl

\bibitem{arai2002advances}
T.~Arai, E.~Pagello, L.~E. Parker \emph{et~al.}, ``Advances in multi-robot
  systems,'' \emph{IEEE Transactions on Robotics and Automation}, vol.~18,
  no.~5, pp. 655--661, 2002.

\bibitem{toth2002overview}
P.~Toth and D.~Vigo, ``An overview of vehicle routing problems,'' \emph{The
  Vehicle Routing Problem}, pp. 1--26, 2002.

\bibitem{cliff2015online}
O.~M. Cliff, R.~Fitch, S.~Sukkarieh, D.~L. Saunders, and R.~Heinsohn, ``Online
  localization of radio-tagged wildlife with an autonomous aerial robot
  system,'' in \emph{Robotics: Science and Systems}, 2015.

\bibitem{fink2008multi}
J.~Fink, M.~A. Hsieh, and V.~Kumar, ``Multi-robot manipulation via caging in
  environments with obstacles,'' in \emph{2008 IEEE International Conference on
  Robotics and Automation}.\hskip 1em plus 0.5em minus 0.4em\relax IEEE, 2008,
  pp. 1471--1476.

\bibitem{varava2017herding}
A.~Varava, K.~Hang, D.~Kragic, and F.~T. Pokorny, ``Herding by caging: a
  topological approach towards guiding moving agents via mobile robots.'' in
  \emph{Robotics: Science and Systems}, 2017, pp. 696--700.

\bibitem{baier2008principles}
C.~Baier and J.-P. Katoen, \emph{Principles of Model Checking}.\hskip 1em plus
  0.5em minus 0.4em\relax MIT press, 2008.

\bibitem{ulusoy2013optimality}
A.~Ulusoy, S.~L. Smith, X.~C. Ding, C.~Belta, and D.~Rus, ``Optimality and
  robustness in multi-robot path planning with temporal logic constraints,''
  \emph{The International Journal of Robotics Research}, vol.~32, no.~8, pp.
  889--911, 2013.

\bibitem{kantaros2020stylus}
Y.~Kantaros and M.~M. Zavlanos, ``Stylus*: A temporal logic optimal control
  synthesis algorithm for large-scale multi-robot systems,'' \emph{The
  International Journal of Robotics Research}, vol.~39, no.~7, pp. 812--836,
  2020.

\bibitem{schillinger2018simultaneous}
P.~Schillinger, M.~B{\"u}rger, and D.~V. Dimarogonas, ``Simultaneous task
  allocation and planning for temporal logic goals in heterogeneous multi-robot
  systems,'' \emph{The International Journal of Robotics Research}, vol.~37,
  no.~7, pp. 818--838, 2018.

\bibitem{guo2018multirobot}
M.~Guo and M.~M. Zavlanos, ``Multirobot data gathering under buffer constraints
  and intermittent communication,'' \emph{IEEE Transactions on Robotics},
  vol.~34, no.~4, pp. 1082--1097, 2018.

\bibitem{guo2015multi}
M.~Guo and D.~V. Dimarogonas, ``Multi-agent plan reconfiguration under local
  ltl specifications,'' \emph{The International Journal of Robotics Research},
  vol.~34, no.~2, pp. 218--235, 2015.

\bibitem{tumova2016multi}
J.~Tumova and D.~V. Dimarogonas, ``Multi-agent planning under local ltl
  specifications and event-based synchronization,'' \emph{Automatica}, vol.~70,
  pp. 239--248, 2016.

\bibitem{guo2016task}
M.~Guo and D.~V. Dimarogonas, ``Task and motion coordination for heterogeneous
  multiagent systems with loosely coupled local tasks,'' \emph{IEEE
  Transactions on Automation Science and Engineering}, vol.~14, no.~2, pp.
  797--808, 2016.

\bibitem{luo2021abstraction}
X.~Luo, Y.~Kantaros, and M.~M. Zavlanos, ``An abstraction-free method for
  multirobot temporal logic optimal control synthesis,'' \emph{IEEE
  Transactions on Robotics}, 2021.

\bibitem{luo2021temporal}
X.~Luo and M.~M. Zavlanos, ``Temporal logic task allocation in heterogeneous
  multi-robot systems,'' \emph{IEEE Transactions on Robotics}, vol.~38, no.~6,
  pp. 3602--3621, 2022.

\bibitem{sahin2019multirobot}
Y.~E. Sahin, P.~Nilsson, and N.~Ozay, ``Multirobot coordination with counting
  temporal logics,'' \emph{IEEE Transactions on Robotics}, vol.~36, no.~4, pp.
  1189--1206, 2019.

\bibitem{jones2019scratchs}
A.~M. Jones, K.~Leahy, C.~Vasile, S.~Sadraddini, Z.~Serlin, R.~Tron, and
  C.~Belta, ``Scratchs: Scalable and robust algorithms for task-based
  coordination from high-level specifications,'' in \emph{Proc. Int. Symp.
  Robot. Res.}, 2019, pp. 1--16.

\bibitem{LAVAEI2022110617}
A.~Lavaei, S.~Soudjani, A.~Abate, and M.~Zamani, ``Automated verification and
  synthesis of stochastic hybrid systems: A survey,'' \emph{Automatica}, vol.
  146, p. 110617, 2022.

\bibitem{torreno2017cooperative}
A.~Torre{\~n}o, E.~Onaindia, A.~Komenda, and M.~{\v{S}}tolba, ``Cooperative
  multi-agent planning: A survey,'' \emph{ACM Computing Surveys (CSUR)},
  vol.~50, no.~6, pp. 1--32, 2017.

\bibitem{gini2017multi}
M.~Gini, ``Multi-robot allocation of tasks with temporal and ordering
  constraints,'' in \emph{AAAI Conference on Artificial Intelligence}, 2017.

\bibitem{khamis2015multi}
A.~Khamis, A.~Hussein, and A.~Elmogy, ``Multi-robot task allocation: A review
  of the state-of-the-art,'' \emph{Cooperative Robots and Sensor Networks
  2015}, pp. 31--51, 2015.

\bibitem{luo2015distributed}
L.~Luo, N.~Chakraborty, and K.~Sycara, ``Distributed algorithms for multirobot
  task assignment with task deadline constraints,'' \emph{IEEE Transactions on
  Automation Science and Engineering}, vol.~12, no.~3, pp. 876--888, 2015.

\bibitem{fukasawa2006robust}
R.~Fukasawa, H.~Longo, J.~Lysgaard, M.~P. De~Arag{\~a}o, M.~Reis, E.~Uchoa, and
  R.~F. Werneck, ``Robust branch-and-cut-and-price for the capacitated vehicle
  routing problem,'' \emph{Mathematical Programming}, vol. 106, no.~3, pp.
  491--511, 2006.

\bibitem{boerkoel2013distributed}
J.~C. Boerkoel~Jr, L.~R. Planken, R.~J. Wilcox, and J.~A. Shah, ``Distributed
  algorithms for incrementally maintaining multiagent simple temporal
  networks,'' in \emph{International Conference on Automated Planning and
  Scheduling}, 2013.

\bibitem{nunes2015multi}
E.~Nunes and M.~Gini, ``Multi-robot auctions for allocation of tasks with
  temporal constraints,'' in \emph{Proceedings of the AAAI Conference on
  Artificial Intelligence}, vol.~29, no.~1, 2015.

\bibitem{FANG2022110228}
J.~Fang, Z.~Zhang, and R.~V. Cowlagi, ``Decentralized route-planning for
  multi-vehicle teams to satisfy a subclass of linear temporal logic
  specifications,'' \emph{Automatica}, vol. 140, p. 110228, 2022.

\bibitem{hoos2004stochastic}
H.~H. Hoos and T.~St{\"u}tzle, \emph{Stochastic Local Search: Foundations and
  Applications}.\hskip 1em plus 0.5em minus 0.4em\relax Elsevier, 2004.

\bibitem{lavalle2006planning}
S.~M. LaValle, \emph{Planning Algorithms}.\hskip 1em plus 0.5em minus
  0.4em\relax Cambridge univ. press, 2006.

\bibitem{chen2005formation}
Y.~Q. Chen and Z.~Wang, ``Formation control: a review and a new
  consideration,'' in \emph{2005 IEEE/RSJ International Conference on
  Intelligent Robots and Systems}.\hskip 1em plus 0.5em minus 0.4em\relax IEEE,
  2005, pp. 3181--3186.

\bibitem{li2009consensus}
Z.~Li, Z.~Duan, G.~Chen, and L.~Huang, ``Consensus of multiagent systems and
  synchronization of complex networks: A unified viewpoint,'' \emph{IEEE
  Transactions on Circuits and Systems I: Regular Papers}, vol.~57, no.~1, pp.
  213--224, 2009.

\bibitem{mesbahi2010graph}
M.~Mesbahi and M.~Egerstedt, \emph{Graph theoretic methods in multiagent
  networks}.\hskip 1em plus 0.5em minus 0.4em\relax Princeton University Press,
  2010, vol.~33.

\bibitem{lahijanian2011temporal}
M.~Lahijanian, S.~B. Andersson, and C.~Belta, ``Temporal logic motion planning
  and control with probabilistic satisfaction guarantees,'' \emph{IEEE
  Transactions on Robotics}, vol.~28, no.~2, pp. 396--409, 2011.

\bibitem{chen2011formal}
Y.~Chen, X.~C. Ding, A.~Stefanescu, and C.~Belta, ``Formal approach to the
  deployment of distributed robotic teams,'' \emph{IEEE Transactions on
  Robotics}, vol.~28, no.~1, pp. 158--171, 2011.

\bibitem{lawler1966branch}
E.~L. Lawler and D.~E. Wood, ``Branch-and-bound methods: A survey,''
  \emph{Operations Research}, vol.~14, no.~4, pp. 699--719, 1966.

\bibitem{morrison2016branch}
D.~R. Morrison, S.~H. Jacobson, J.~J. Sauppe, and E.~C. Sewell,
  ``Branch-and-bound algorithms: A survey of recent advances in searching,
  branching, and pruning,'' \emph{Discrete Optimization}, vol.~19, pp. 79--102,
  2016.

\bibitem{junger1995traveling}
M.~J{\"u}nger, G.~Reinelt, and G.~Rinaldi, ``The traveling salesman problem,''
  \emph{Handbooks in Operations Research and Management Science}, vol.~7, pp.
  225--330, 1995.

\bibitem{brucker1994branch}
P.~Brucker, B.~Jurisch, and B.~Sievers, ``A branch and bound algorithm for the
  job-shop scheduling problem,'' \emph{Discrete Applied Mathematics}, vol.~49,
  no. 1-3, pp. 107--127, 1994.

\bibitem{simovici2008mathematical}
D.~A. Simovici and C.~Djeraba, ``Mathematical tools for data mining,''
  \emph{SpringerVerlag, London}, 2008.

\bibitem{pan2007multi}
X.~Pan, C.~S. Han, K.~Dauber, and K.~H. Law, ``A multi-agent based framework
  for the simulation of human and social behaviors during emergency
  evacuations,'' \emph{AI \& Society}, vol.~22, no.~2, pp. 113--132, 2007.

\bibitem{hochba1997approximation}
D.~S. Hochba, ``Approximation algorithms for np-hard problems,'' \emph{ACM
  Sigact News}, vol.~28, no.~2, pp. 40--52, 1997.

\bibitem{bovet1994introduction}
D.~P. Bovet, P.~Crescenzi, and D.~Bovet, \emph{Introduction to the Theory of
  Complexity}.\hskip 1em plus 0.5em minus 0.4em\relax Prentice Hall London,
  1994, vol.~7.

\bibitem{gastin2001fast}
P.~Gastin and D.~Oddoux, ``Fast {LTL} to b{\"u}chi automata translation,'' in
  \emph{Computer Aided Verification}.\hskip 1em plus 0.5em minus 0.4em\relax
  Springer, 2001, pp. 53--65.

\bibitem{sedgewick2001algorithms}
R.~Sedgewick, \emph{Algorithms in C, part 5: graph algorithms}.\hskip 1em plus
  0.5em minus 0.4em\relax Pearson Education, 2001.

\bibitem{lima2010ibm}
R.~Lima and E.~Seminar, ``Ibm ilog cplex-what is inside of the box,'' in
  \emph{Proc. 2010 EWO Seminar}, 2010, pp. 1--72.

\bibitem{makhorin2008glpk}
A.~Makhorin, ``Glpk (gnu linear programming kit),'' \emph{http://www. gnu.
  org/s/glpk/glpk. html}, 2008.

\end{thebibliography}
\end{document}